\documentclass[a4paper,11pt]{article}
\usepackage[utf8]{inputenc}
\usepackage[T1]{fontenc}

\usepackage{verbatim} 

\usepackage{amssymb,amsthm,amsfonts,stmaryrd,amsmath,algorithm,algorithmicx,bbm,eucal,placeins,graphicx}
\usepackage{subfigure}
\usepackage[top=3cm,bottom=3cm,left=3cm,right=3cm]{geometry}
\usepackage[authoryear,round]{natbib}
\usepackage[pagebackref = true]{hyperref}
\hypersetup{
  colorlinks = true,
  urlcolor = blue, 
  linkcolor = blue,
  citecolor = blue,
  pdftitle = {Sequential Learning of Principal Curves: Summarizing Data Streams on the Fly - \today},
  pdfauthor = {Benjamin Guedj and Le Li},
  pdfsubject = {}
}

\newtheorem{theo}{Theorem}

\newtheorem{coro}{Corollary}
\newtheorem{lem}{Lemma}

\def\R{\mathbb{R}}
\def\E{\mathbb{E}}

\def\PP{\mathcal{P}}
\def\K{\mathcal{K}}

\def\cball{B(\text{0}, \sqrt{d}R)}
\def\f{\mathbf{f}}
\def\grid{\mathcal{Q}_{\delta}}
\def\PP{\mathcal{P}}

\def\Vf{\overset{\rightharpoonup}{\mathbf{V}}}
\def\e{\mathrm{e}}
\def\farginf{\underset{\f \in \mathcal{F}}{\arg\inf}}
\def\At{\mathcal{A}_{t}}

\def\cumrwrd{\sum_{t=1}^{T}\E\left[r_{\hat{\f}_{t}, t}\right]}

\def\cumrwrd{\sum_{t=1}^{T}\E\left[r_{\hat{\f}_{t}, t}\right]}

\def\condilocrwrd{\E\left[r_{\hat{\f}_{t}, t}\bigg|\mathcal{H}_{t}\right]}

\def\exphatordonwhole{\E\left[\hat{r}_{\hat{\sigma}^{t}\left(\mathcal{A}_{t}\right), t}\right]}

\def\ordonwhole{r_{\hat{\sigma}^{t}\left(\mathcal{F}_{p}\right), t}}
\def\expglbhatordonwholepenalty{\E\left[\max_{\hat{\sigma}}\left\{\sum_{t=1}^{T}\hat{r}_{\hat{\sigma}\left(\mathcal{A}_{t}\right), t} - \frac{1}{\eta}h\left(\hat{\sigma}\left(\mathcal{A}_{t}\right)\right)\right\}\right]}
\def\expglbordonwholepenalty{\E\left[\max_{\sigma}\left\{\sum_{t=1}^{T}r_{{\sigma}\left(\mathcal{A}_{t}\right), t} - \frac{1}{\eta}h\left(\sigma\left(\mathcal{A}_{
t}\right)\right)\right\}\right]}
\def\glbhatordonwholepenalty{\max_{\hat{\sigma}}\left\{\sum_{t=1}^{T}\hat{r}_{\hat{\sigma}\left(\mathcal{A}_{t}\right), t} - \frac{1}{\eta}h\left(\hat{\sigma}\left(\mathcal{A}_{
t}\right)\right)\right\}}
\def\glbordonwholepenalty{\max_{\sigma}\left\{\sum_{t=1}^{T}r_{{\sigma}\left(\mathcal{A}_{t}\right), t} - \frac{1}{\eta}h\left(\sigma\left(\mathcal{A}_{
t}\right)\right)\right\}}

\DeclareMathOperator*{\argmax}{arg\,max}
\DeclareMathOperator*{\argmin}{arg\,min}

\renewcommand{\P}{\mathbb{P}}

\title{Sequential Learning of Principal Curves: Summarizing Data Streams on the Fly}
\author{Benjamin \textsc{Guedj}\footnote{Inria and University College London -- corresponding author, \href{mailto:benjamin.guedj@inria.fr}{benjamin.guedj@inria.fr}} ~and Le \textsc{Li}\footnote{Université d'Angers and iAdvize}}

\begin{document}

\maketitle



\begin{abstract}
When confronted with massive data streams, summarizing data with dimension reduction methods such as PCA raises theoretical and algorithmic pitfalls. Principal curves act as a nonlinear generalization of PCA and the present paper proposes a novel algorithm to automatically and sequentially learn principal curves from data streams. We show that our procedure is supported by regret bounds with optimal sublinear remainder terms. A greedy local search implementation (called \texttt{slpc}, for Sequential Learning Principal Curves) that incorporates both sleeping experts and multi-armed bandit ingredients is presented, along with its regret computation and performance on synthetic and real-life data.
\end{abstract}





\section{Introduction}

\begin{figure}[h]
\vspace{-2.5cm}
\centering
\includegraphics[clip,width = 5cm, angle = -40]{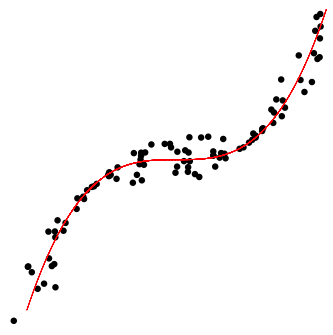}
\vspace{-2.5cm}
\caption{A principal curve.}
\label{fig1}
\end{figure}

Numerous methods have been proposed in the statistics and machine learning literature to sum up information and represent data by condensed and simpler-to-understand quantities.
Among those methods, Principal Component Analysis (PCA) aims at identifying the maximal variance axes of data. This serves as a way to represent data in a more compact fashion and hopefully reveal as well as possible their variability. PCA has been introduced by \cite{peason1901lines} and \cite{spearman1904general} and further developed by \cite{hotelling1933analysis}. This is one of the most widely used procedures in multivariate exploratory analysis targeting dimension reduction or features extraction.
Nonetheless, PCA is a linear procedure and the need for more sophisticated nonlinear techniques has led to the notion of principal curve. Principal curves may be seen as a nonlinear generalization of the first principal component. The goal is to obtain a curve which passes "in the middle" of data, as illustrated by \autoref{fig1}. This notion of skeletonization of data clouds has been at the heart of numerous applications in many different domains, such as physics \citep{FO1989,BRU2007}, character and speech recognition \citep{RN1999,KK2002}, mapping and geology \citep{banfield1992ice,stanford2000finding,BRU2007}, to name but a few.

\subsection{Earlier works on principal curves}

\begin{figure}[h]
\vspace{-2.5cm}
\centering
\includegraphics[width = 6cm, height = 5cm, angle = -40]{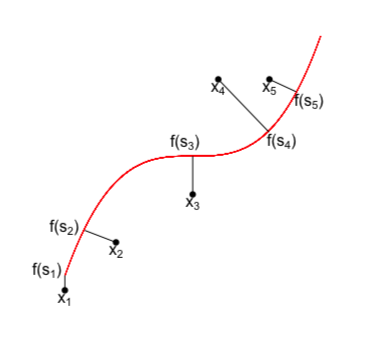}
\vspace{-3cm}
\caption{A principal curve and projections of data onto it.}
\label{fig:fig2}
\end{figure}

The original definition of principal curve dates back to \cite{HS1989}. A principal curve is a smooth ($C^{\infty}$) parameterized curve $\mathbf{f}(s) = \left(f_{1}(s), \dots, f_{d}(s)\right)$ in $\R^{d}$ which does not intersect itself, has finite length inside any bounded subset of $\mathbb{R}^{d}$ and is self-consistent. This last requirement means that $\mathbf{f}(s) = \mathbb{E}[X\vert s_{\mathbf{f}}(X) = s]$,
where $X \in \mathbb{R}^{d}$ is a random vector and the so-called projection index $s_{\mathbf{f}}(x)$ is the largest real number $s$ minimizing the squared Euclidean distance between $\mathbf{f}(s)$ and $x$, defined by
\begin{equation*}
s_{\f}(x) = \sup\left\{s: \left\|x-\f(s)\right\|_{2}^2 = \inf_{\tau}\ \left\|x-\f(\tau)\right\|_{2}^2\right\}.
\end{equation*}
Self-consistency means that each point of $\mathbf{f}$ is the average (under the distribution of $X$) of all data points projected on $\mathbf{f}$, as illustrated by \autoref{fig:fig2}.
However, an unfortunate consequence of this definition is that the existence is not guaranteed in general for a particular distribution, let alone for an online sequence for which no probabilistic assumption is made. \cite{KEG1999} proposed a new concept of principal curves which ensures its existence for a large class of distributions. Principal curves $\f^{\star}$ are defined as the curves minimizing the expected squared distance over a class $\mathcal{F}_{L}$ of curves whose length is smaller than $L>0$, namely, 
\begin{equation*}
   \f^{\star} \in \underset{\f \in \mathcal{F}_{L}}{\arg\inf} \ \Delta(\f), 
\end{equation*} 
where
\begin{equation*}
    \Delta(\f) = \E\left[\Delta\left(\f,X\right)\right] = \E\left[\inf_{s}\ \left\|\f(s)-X\right\|_2^2\right].
\end{equation*}
If $\E\|X\|_2^2 < \infty$, $\f^{\star}$ always exists but may not be unique. In practical situation where only i.i.d copies $X_{1}, \dots, X_{n}$ of $X$ are observed,   \cite{KEG1999} considers classes $\mathcal{F}_{k,L}$ of all polygonal lines with $k$ segments and length not exceeding $L$, and chooses an estimator $\hat{\f}_{k,n}$ of $\f^{\star}$ as the one within $\mathcal{F}_{k,L}$ which minimizes the empirical counterpart
\begin{equation*}
    \Delta_{n}(\f) = \frac{1}{n}\sum_{i=1}^{n}\Delta\left(\f, X_{i}\right)
\end{equation*}
of $\Delta(\f)$. It is proved in \cite{KKLZ2000} that if $X$ is almost surely bounded and $k\propto n^{1/3}$, then
\begin{equation*}
    \Delta\left(\hat{\f}_{k,n}\right) - \Delta\left(\f^{\star}\right) = \mathcal{O}\left(n^{-1/3}\right).
\end{equation*}
As the task of finding a polygonal line with $k$ segments and length at most $L$ that minimizes $\Delta_{n}(\f)$ is computationally costly, \cite{KKLZ2000} proposes the Polygonal Line algorithm. This iterative algorithm proceeds by fitting a polygonal line with $k$ segments and considerably speeds up the exploration part by resorting to gradient descent. The two steps (projection and optimization) are similar to what is done by the $k$-means algorithm. However, the Polygonal Line algorithm is not supported by theoretical bounds and leads to variable performance depending on the distribution of the observations.
\medskip
 
\begin{figure}[ht!]
\centering
\subfigure[A too small $k$.]{\includegraphics[width=4.6cm, angle =0]{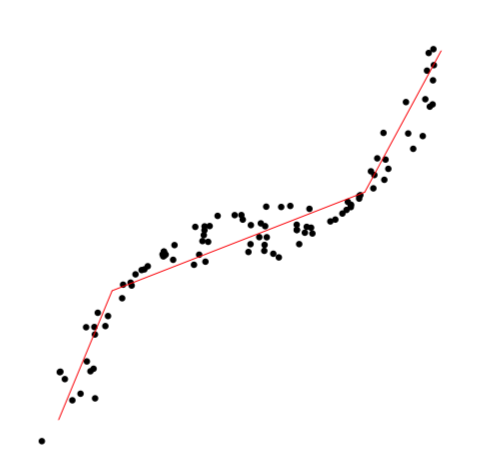}}
\hfill
\subfigure[Right $k$.]{\includegraphics[width=4.6cm, angle = 0]{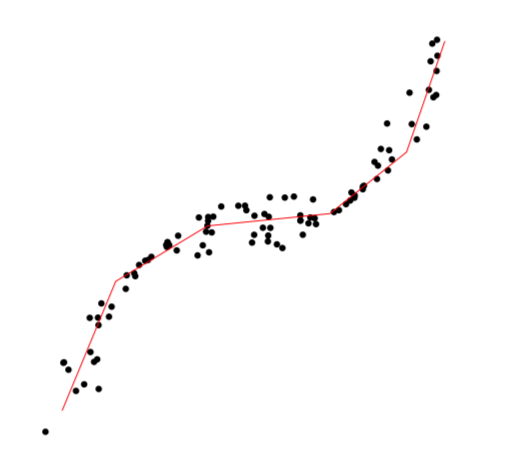}}
\hfill
\subfigure[A too large $k$.]{\includegraphics[width=4.6cm, angle = 0]{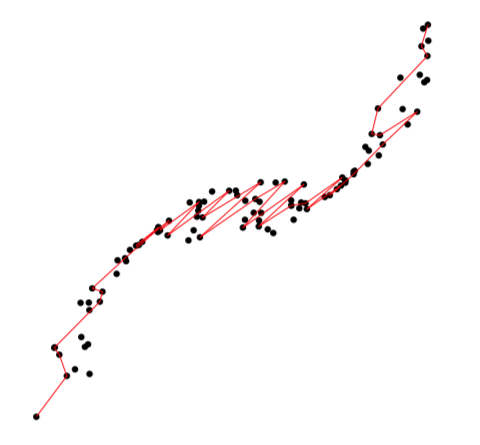}}
\caption{Principal curves with different number $k$ of segments.}
\label{fig:fig3}
\end{figure}

As the number $k$ of segments plays a crucial role (a too small $k$ leads to a poor summary of data while a too large $k$ yields overfitting, see \autoref{fig:fig3}), \cite{BA2012} aim to fill the gap by selecting an optimal $k$ from both theoretical and practical perspectives.
Their approach relies strongly on the theory of model selection by penalization introduced by \cite{BBM1999} and further developed by \cite{BM2007}. By  considering countable classes $\{\mathcal{F}_{k,\ell}\}_{k,\ell}$ of polygonal lines with $k$ segments, total length $\ell\leq L$ and whose vertices are on a lattice, the optimal $(\hat{k}, \hat{\ell})$ is obtained as the minimizer of the criterion
\begin{equation*}
    \text{crit}(k,\ell) = \Delta_{n}\left(\hat{\f}_{k,\ell}\right) + \text{pen}(k,\ell),
\end{equation*}
where 
\begin{equation*}
    \text{pen}(k,\ell) = c_{0}\sqrt{\frac{k}{n}} + c_{1}\frac{\ell}{n} + c_{2}\frac{1}{\sqrt{n}} + \delta^2\sqrt{\frac{w_{k,\ell}}{2n}}
\end{equation*}
is a penalty function where $\delta$ stands for the diameter of observations, $w_{k,\ell}$ denotes the weight attached to class $\mathcal{F}_{k,\ell}$ and with constants $c_{0},c_{1}, c_{2}$ depending on $\delta$, the maximum length $L$ and the dimension of observations. \cite{BA2012} then prove that
\begin{equation}\label{eq:batch upper bound}
    \mathbb{E}\left[\Delta(\hat{\f}_{\hat{k},\hat{\ell}})-\Delta(\f^{\star})\right]  \leq  \inf_{k,\ell}\ \Big\{\mathbb{E}\left[\Delta(\hat{\f}_{k,\ell})-\Delta(\f^{\star})\right] + \text{pen}(k,\ell)\Big\} + \frac{\delta^2\Sigma}{2^{3/2}}\sqrt{\frac{\pi}{n}},
\end{equation}
where $\Sigma$ is a numerical constant. The expected loss of the final polygonal line $\hat{\f}_{\hat{k},\hat{\ell}}$ is close to the minimal loss achievable over $\mathcal{F}_{k,\ell}$ up to a remainder term decaying as $1/\sqrt{n}$.

\subsection{Motivation}

The big data paradigm---where collecting, storing and analyzing massive amounts of large and complex data becomes the new standard---commands to revisit some of the classical statistical and machine learning techniques. The tremendous improvements of data acquisition infrastructures generates new continuous streams of data, rather than batch datasets. This has drawn a large interest to sequential learning. Extending the notion of principal curves to the sequential settings opens immediate practical application possibilities. As an example, path planning for passengers' location can help taxi companies to better optimize their fleet. Online algorithms that could yield instantaneous path summarization would be adapted to the sequential nature of geolocalized data. Existing theoretical works and practical implementations of principal curves are designed for the batch setting \citep{KEG1999,KKLZ2000,KK2002,SK2002,BA2012} and their adaptation to the sequential setting is not a smooth process. As an example, consider the algorithm in \cite{BA2012}. It is assumed that vertices of principal curves are located on a lattice, and its computational complexity is of order $\mathcal{O}(nN^{p})$ where $n$ is the number of observations, $N$ the number of points on the lattice and $p$ the maximum number of vertices. When $p$ is large, running this algorithm at each epoch yields a monumental computational cost. In general, if data is not identically distributed or even adversary, algorithms that originally worked well in the batch setting may not be ideal when cast onto the online setting \citep[see][Chapter 4]{CBL2006}.  To the best of our knowledge, very little effort has been put so far into extending principal curves algorithms to the sequential context \citep[to the notable exception of][in a fairly different setting and with no theoretical results]{laparra2016sequential}. The present paper aims at filling this gap: our goal is to propose an online perspective to principal curves by automatically and sequentially learning the best principal curve summarizing a data stream. 
Sequential learning takes advantage of the latest collected (set of) observations and therefore suffers a much smaller computational cost.
\medskip

Sequential learning operates as follows: a blackbox reveals at each time $t$ some deterministic value $x_t, t= 1,2,\dots$, and a forecaster attempts to predict sequentially the next value based on past observations (and possibly other available information). The performance of the forecaster is no longer evaluated by its generalization error (as in the batch setting) but rather by a regret bound which quantifies the cumulative loss of a forecaster in the first $T$ rounds with respect to some reference minimal loss. In sequential learning, the velocity of algorithms may be favored over statistical precision. An immediate use of aforecited techniques \citep{KKLZ2000,SK2002,BA2012} at each time round $t$ (treating data collected until $t$ as a batch dataset) would result in a monumental algorithmic cost. Rather, we propose a novel algorithm which adapts to the sequential nature of data, \emph{i.e.}, which takes advantage of previous computations.
\medskip

The contributions of the present paper are twofold. We first propose a sequential principal curves algorithm, for which we derive regret bounds. We then move towards an implementation, illustrated on a toy dataset and a real-life dataset (seismic data). The sketch of our algorithm procedure is as follows. At each time round $t$, the number of segments of $k_{t}$ is chosen automatically and the number of segments $k_{t+1}$ in the next round is obtained by only using information about $k_{t}$ and a small amount of past observations. The core of our procedure relies on computing a quantity which is linked to the mode of the so-called Gibbs quasi-posterior and is inspired by quasi-Bayesian learning. The use of quasi-Bayesian estimators is especially advocated by the PAC-Bayesian theory which originates in the machine learning community in the late 1990s, in the seminal works of \cite{STW1997} and \citet{McA1998,McA1999}. The PAC-Bayesian theory has been successfully adapted to sequential learning problems, see for example \cite{LGL2017} for online clustering.
\medskip

The paper is organized as follows. \autoref{sec:notation} presents our notation and our online principal curve algorithm, for which we provide regret bounds with sublinear remainder terms in \autoref{sec:regret bounds}. A practical implementation is proposed in \autoref{sec:implementation} and we illustrate its performance on synthetic and real-life data sets in \autoref{sec:simus}. Proofs to all original results claimed in the paper are collected in \autoref{sec:proofs}.

\section{Notation}\label{sec:notation}
\begin{figure}[h]
\centering
\includegraphics[scale=.4]{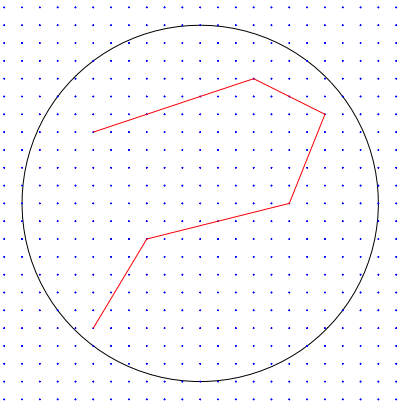}
\caption{An example of a lattice $\Gamma_{\delta}$ in $\R^2$ with $\delta = 1$ (spacing between blue points) and $B(0, 10)$ (black circle). The red polygonal line is composed with vertices in $\mathcal{Q}_{\delta}=B(0,10)\cap \Gamma_{\delta}$. }
\label{fig:fig4}
\end{figure}

A parameterized curve in $\R^d$ is a continuous function $\f : I \longrightarrow \R^d$ where $I = [a,b]$ is a closed interval of the real line. The length of $\f$ is given by 
\begin{equation*}
\mathcal{L}(\f) =  \lim_{M\to\infty} \ \left\{ \underset{a = s_{0} < s_{1} < \dots < s_{M} = b}{\sup}\ \  \sum_{i=1}^{M}\|\f(s_{i}) - \f(s_{i-1})\|_{2}\right\}.
\end{equation*}
Let $x_{1}, x_{2}, \dots, x_{T} \in B(0, \sqrt{d}R) \subset \R^d$ be a sequence of data, where $B(\mathbf{c},R)$ stands for the $\ell_{2}$-ball centered in $\mathbf{c} \in \R^d$ with radius $R>0$. Let $\mathcal{Q}_{\delta}$ be a grid over $B(0, \sqrt{d}R)$, \emph{i.e.}, $\mathcal{Q}_{\delta} = \cball \cap \Gamma_{\delta}$ where $\Gamma_{\delta}$ is a lattice in $\R^d$ with spacing $\delta >0$.
Let $L>0$ and define for each $k \in \llbracket1,p\rrbracket$ the collection $\mathcal{F}_{k,L}$ of polygonal lines $\f$ with $k$ segments whose vertices are in $\mathcal{Q}_{\delta}$ and such that $\mathcal{L}(\f) \leq L$.
Denote by $\mathcal{F}_{p} = \cup_{k =1}^{p} \mathcal{F}_{k,L}$ all polygonal lines with a number of segments $\leq p$, whose vertices are in $\grid$ and whose length is at most $L$. Finally, let $\mathcal{K}(\f)$ denote the number of segments of $\f \in \mathcal{F}_{p}$. This strategy is illustrated by  \autoref{fig:fig4}.

Our goal is to learn a time-dependent polygonal line which passes through the "middle" of data and gives a summary of all available observations $x_{1},\dots, x_{t-1}$ (denoted by $(x_{s})_{1:(t-1)}$ hereafter) before time $t$. Our output at time $t$ is a polygonal line $\hat{\f}_{t} \in \mathcal{F}_{p}$ depending on past information $(x_{s})_{1:(t-1)}$ and past predictions $(\hat{\f}_{s})_{1:(t-1)}$. When $x_{t}$ is revealed, the instantaneous loss at time $t$ is computed as
\begin{equation}\label{def:loss}
\Delta\left(\hat{\f}_{t},x_{t}\right)=\inf_{s\in I}\ \|\hat{\f}_{t}(s)-x_{t}\|^{2}_{2}.
\end{equation}
In what follows, we investigate regret bounds for the cumulative loss based on \eqref{def:loss}. Given a measurable space $\Theta$ (embedded with its Borel $\sigma$-algebra), we let $\PP(\Theta)$ denote the set of probability distributions on $\Theta$, and for some reference measure $\pi$, we let $\PP_\pi(\Theta)$ be the set of probability distributions absolutely continuous with respect to $\pi$.
\medskip

For any $k \in \llbracket1,p\rrbracket $, let $\pi_{k}$ denote a probability distribution on $\mathcal{F}_{k, L}$. We define the \emph{prior} $\pi$ on $\mathcal{F}_{p}=\cup_{k = 1}^{p}\mathcal{F}_{k, L}$ as
\begin{equation*}
\pi(\f)=\sum_{k \in \llbracket1,p\rrbracket }w_{k}\pi_{k}(\f)\mathbbm{1}_{\left\{\f \in \mathcal{F}_{k, L}\right\}},\quad \f\in\mathcal{F}_{p},
\end{equation*}
where $w_1,\dots,w_p\geq 0$ and $\sum_{k \in \llbracket1,p\rrbracket }w_{k} = 1$.
\medskip

We adopt a quasi-Bayesian-flavored procedure: consider the Gibbs quasi-posterior (note that this is not a proper posterior in all generality, hence the term "quasi")
\begin{equation*}
    \hat{\rho}_t(\cdot) \propto \exp(-\lambda S_t(\cdot))\pi(\cdot),
\end{equation*}
where 
\begin{equation*}
S_{t}(\f)=S_{t-1}(\f)+\Delta(\f,x_{t})+\frac{\lambda}{2}\big(\Delta(\f,x_{t})-\Delta(\hat{\f}_{t},x_{t})\big)^2,
\end{equation*}
as advocated by \cite{Aud2009} and \cite{LGL2017} who then consider realisations from this quasi-posterior. In the present paper, we will rather focus on a quantity linked to the mode of this quasi-posterior. Indeed, the mode of the quasi-posterior $\hat{\rho}_{t+1}$ is 
\begin{equation*}
\arg\min_{\f \in \mathcal{F}_{p}}\Bigg\{\underbrace{\sum_{s=1}^{t}\Delta(\f, x_{s})}_{(i)} + \underbrace{\frac{\lambda}{2} \sum_{s=1}^{t}\big(\Delta(\f,x_{t})-\Delta(\hat{\f}_{t},x_{t})\big)^2}_{(ii)}   + \underbrace{\frac{\ln\pi(\f)}{\lambda}}_{(iii)}\Bigg\},
\end{equation*} 
where \emph{(i)} is a cumulative loss term, \emph{(ii)} is a term controlling the variance of the prediction $\f$ to past predictions $\hat{\f}_{s}, s\leq t$, and \emph{(iii)} can be regarded as a penalty function on the complexity of $\f$ if $\pi$ is well chosen. This mode hence has a similar flavor to follow the best expert or follow the perturbed leader in the setting of prediction with experts (see \citealp{HP2005} and \citealp[][Chapters 3 and 4]{CBL2006}) if we consider each $\f \in \mathcal{F}_{p}$ as an expert which always delivers constant advice. These remarks yield \autoref{proc:procedure 1}. 


\begin{algorithm}[ht]
\caption{Sequentially learning principal curves}
\label{proc:procedure 1}
\begin{algorithmic}[1]
\\\textbf{Input parameters}: $p>0, \eta>0, \pi(z) = \e^{-z}\mathbbm{1}_{\{z >0 \}}$ and penalty function $h:\mathcal{F}_{p} \to \R^{+}$
\\\textbf{Initialization}: For each $\f \in \mathcal{F}_{p}$, draw $z_{\f}  \sim \pi$ and $\Delta_{\f, 0} = \frac{1}{\eta}(h(\f) - z_{\f})$
\State \textbf{For $t=1,\dots,T$}
\State\hspace{\algorithmicindent} Get the data $x_{t}$
\State \hspace{\algorithmicindent} Obtain
\begin{equation*}
\hat{\f}_{t} =  \underset{\f \in \mathcal{F}_{p}}{\arg\inf}\left\{\sum_{s=0}^{t-1}\Delta_{\f, s}\right\},
\end{equation*}
 \hspace{\algorithmicindent} where $\Delta_{\f,s} = \Delta(\f, x_{s})$, $s\geq 1$.
\State \textbf{End for}
\end{algorithmic}
\end{algorithm}

\section{Regret bounds for sequential learning of principal curves}\label{sec:regret bounds}

We now present our main theoretical results.
\begin{theo}\label{thm:theorem 1}
For any sequence $(x_{t})_{1:T} \in \cball$, $R\geq 0$ and any penalty function $h:\mathcal{F}_{p} \to \R^{+}$, let $\pi(z) = \e^{-z}\mathbbm{1}_{\{z >0 \}}$. Let $ 0<\eta \leq \frac{1}{d(2R+\delta)^2}$, then the procedure described in  \autoref{proc:procedure 1} satisfies
\begin{equation*}
\sum_{t=1}^T\mathbb{E}
_{\pi}\left[\Delta(\hat{\f}_{t},x_{t})\right] \leq  \left(1+c_{0}(\e -1)\eta \right)S_{T, h, \eta} + \frac{1}{\eta}\left(1+\ln \sum_{\f \in \mathcal{F}_{p}}\e^{-h(\f)}\right),
\end{equation*}
where $c_{0} = d(2R+\delta)^{2}$ and 
\begin{equation*}
 S_{T, h, \eta} = \inf_{k \in \llbracket1,p\rrbracket}\left\{\inf_{\substack{\f \in \mathcal{F}_{p} \\ \mathcal{K}(\f) = k}}\left\{\sum_{t=1}^{T}\Delta(\f, x_{t}) + \frac{h(\f)}{\eta}\right\}\right\}.
\end{equation*}
\end{theo}
The expectation of the cumulative loss of polygonal lines $\hat{\f}_{1}, \dots, \hat{\f}_{T}$ is upper-bounded by the smallest penalised cumulative loss over all $k \in \{1,\dots, p\}$ up to a multiplicative term $(1+c_{0}(\e-1)\eta)$ which can be made arbitrarily close to 1 by choosing a small enough $\eta$. However, this will lead to both a large $h(\f)/\eta$ in $S_{T,h,\eta}$ and a large $\frac{1}{\eta}(1+\ln \sum_{\f \in \mathcal{F}_{p}}\e^{-h(\f)})$. In addition, another important issue is the choice of the penalty function $h$. For each $\f \in \mathcal{F}_{p}$, $h(\f)$ should be large enough to ensure a small $\sum_{\f \in \mathcal{F}_{p}}\e^{-h(\f)}$ while not too large to avoid overpenalization and a larger value for $S_{T, h, \eta}$. We therefore set 
\begin{equation}\label{eq:choice for penalty}
   h(\f) \geq \ln(p\e) + \ln \bigg|\{\f \in \mathcal{F}_{p}, \mathcal{K}(\f) = k\}\bigg| \end{equation}
for each $\f$ with $k$ segments (where $|M|$ denotes the cardinality of a set $M$) since
it leads to 
\begin{equation*}
    \sum_{\f \in \mathcal{F}_{p}}\e^{-h(\f)}) = \sum_{k \in \llbracket1,p\rrbracket}\sum_{\substack{\f \in \mathcal{F}_{p} \\ \mathcal{K}(\f) = k}}\e^{-h(\f)}\leq \sum_{k \in \llbracket1,p\rrbracket}\frac{1}{p\e}\leq \frac{1}{\e}.
\end{equation*}
The penalty function $h(\f) = c_{1}\K(\f) + c_{2}L +c_{3}$ satisfies \eqref{eq:choice for penalty}, where $c_{1}, c_{2}, c_{3}$ are constants depending on $R$, $d$, $\delta$, $p$ (this is proven in \autoref{lem:lemma 3}, in \autoref{sec:proofs}). We therefore obtain the following corollary.
\begin{coro}\label{cor:corollary 1}
Under the assumptions of \autoref{thm:theorem 1}, let 
\begin{equation*}
    \eta = \\ \min\left\{\frac{1}{d(2R+\delta)^2}, \sqrt{\frac{c_{1}p + c_{2}L+ c_{3}}{c_{0}(\e-1)\inf_{\f \in \mathcal{F}_{p}} \sum_{t=1}^{T}\Delta(\f,x_{t})}}\right\}.
\end{equation*}
Then
\begin{multline*}
    \sum_{t=1}^{T}\E\left[\Delta(\hat{\f}_{t}, x_{t})\right]  \leq  \inf_{k \in \llbracket1,p\rrbracket}\left\{\inf_{\substack{\f \in \mathcal{F}_{p} \\ \mathcal{K}(\f) = k}}\left\{\sum_{t=1}^{T}\Delta(\f, x_{t}) + \sqrt{c_{0}(\e-1)r_{T,k,L}}\right\}\right\} \\
    + \sqrt{c_{0}(\e-1)r_{T,p,L}} + c_{0}(\e-1)(c_{1}p + c_{2}L + c_{3}),
\end{multline*}
where $r_{T,k,L} = \inf_{\f \in \mathcal{F}_{p}} \sum_{t=1}^{T}\Delta(\f,x_{t})(c_{1}k + c_{2}L + c_{3})$.
\end{coro}
\begin{proof} Note that
\begin{equation*}
    \sum_{t=1}^{T}\E\left[\Delta(\hat{\f}_{t}, x_{t})\right] 
    \leq S_{T, h, \eta}  + \eta c_{0}(\e-1) \inf_{\f \in \mathcal{F}_{p}} \sum_{t=1}^{T}\Delta(\f,x_{t}) +c_{0}(\e-1)(c_{0}p + c_{2}L+ c_{3}),
\end{equation*}
and we conclude by setting $$\eta =  \sqrt{\frac{c_{1}p + c_{2}L+ c_{3}}{c_{0}(\e-1)\inf_{\f \in \mathcal{F}_{p}} \sum_{t=1}^{T}\Delta(\f,x_{t})}}.$$
\end{proof}
Sadly, \autoref{cor:corollary 1} is not of much practical use since the optimal value for $\eta$ depends on $\inf_{\f \in \mathcal{F}_{p}} \sum_{t=1}^{T}\Delta(\f,x_{t})$ which is obviously unknown, even more so at time $t=0$. We therefore provide an adaptive refinement of \autoref{proc:procedure 1} in the following \autoref{proc:procedure 2}.
\begin{algorithm}[ht]
\caption{Sequentially and adaptively learning principal curves}
\label{proc:procedure 2}
\begin{algorithmic}[1]
\\\textbf{Input parameters}: $p>0$, $L>0$, $\pi$, $h$ and $\eta_{0} = \frac{\sqrt{c_1p+c_{2}L+c_3}}{c_{0}\sqrt{\e-1}}$
\\\textbf{Initialization}: For each $\f \in \mathcal{F}_{p}$, draw $z_{\f}  \sim \pi$, $\Delta_{\f, 0} = \frac{1}{\eta_{0}}(h(\f) - z_{\f})$ and $\hat{\f}_{0} = \underset{\f \in \mathcal{F}_{p}}{\mathrm{\arg\inf}}\ \Delta_{\f, 0}$
\State \textbf{For $t=1,\dots,T$}
\State\hspace{\algorithmicindent} Compute $\eta_{t} = \frac{\sqrt{c_1p+c_{2}L+c_3}}{c_{0} \sqrt{(\e-1)t}}$
\State\hspace{\algorithmicindent} Get data $x_{t}$ and compute $\Delta_{\f,t} = \Delta(\f, x_{t}) + \left(\frac{1}{\eta_{t}} - \frac{1}{\eta_{t-1}}\right)\left(h(\f) - z_{\f}\right)$
\State \hspace{\algorithmicindent} Obtain
\begin{equation}\label{eq:cum sum}
\hat{\f}_{t} =  \underset{\f \in \mathcal{F}_{p}}{\mathrm{\arg\inf}}\ \left\{\sum_{s=0}^{t-1}\Delta_{\f, s}\right\}.
\end{equation}
\State \textbf{End for}
\end{algorithmic}
\end{algorithm}
\begin{theo}\label{thm:theorem 2}
For any sequence $(x_{t})_{1:T} \in \cball, R\geq 0$,
let $h(\f) = c_{1}\mathcal{K}(\f)+c_{2}L + c_{3}$ where $c_{1}$, $c_{2}$, $c_{3}$ are constants depending on $R, d, \delta, \ln p$. Let $\pi(z) = \e^{-z}\mathbbm{1}_{\{z >0 \}}$ and $$\eta_{0} = \frac{\sqrt{c_1p+c_{2}L+c_3}}{c_{0}\sqrt{\e-1}}, \quad \eta_{t} = \frac{\sqrt{c_1p+c_{2}L+c_3}}{c_{0} \sqrt{(\e-1)t}}, $$ where $t\geq 1$ and $c_{0}=d(2R+\delta)^2$. Then the procedure described in  \autoref{proc:procedure 2} satisfies
\begin{multline*}
    \sum_{t=1}^{T}\E\left[\Delta(\hat{\f}_{t}, x_{t})\right]
    \leq \inf_{k \in \llbracket1,p\rrbracket}\Bigg\{\inf_{\substack{\f \in \mathcal{F}_{p} \\ \mathcal{K}(\f) = k}}\bigg\{\sum_{t=1}^{T}\Delta(\f, x_{t}) + c_{0}\sqrt{(\e-1)T(c_{1}k+c_2L+c_{3})}\bigg\}\Bigg\} \\
     + 2c_{0}\sqrt{(\e-1)T(c_{1}p+c_2L+c_{3})}.
\end{multline*}
\end{theo}
The message of this regret bound is that the expected cumulative loss of polygonal lines $\hat{\f}_{1}, \dots, \hat{\f}_{T}$ is upper-bounded by the minimal cumulative loss over all $k \in \{1,\dots, p\}$, up to an additive term which is sublinear in $T$. The actual magnitude of this remainder term is $\sqrt{kT}$. When $L$ is fixed, the number $k$ of segments is a measure of complexity of the retained polygonal line. This bound therefore yields the same magnitude than \eqref{eq:batch upper bound} which is the most refined bound in the literature so far \citep[][where the optimal values for $k$ and $L$ are obtained in a model selection fashion]{BA2012}.

\section{Implementation}\label{sec:implementation}

The argument of the infimum in \autoref{proc:procedure 2} is taken over $\mathcal{F}_{p} = \cup_{k=1}^{p}\mathcal{F}_{k, L}$ which has a cardinality of order $\left|\mathcal{Q}_{\delta}\right|^p$, making any greedy search largely time-consuming. We instead turn to the following strategy: given a polygonal line $\hat{\f}_{t} \in \mathcal{F}_{k_{t}, L}$ with $k_{t}$ segments, we consider, with a certain proportion, the availability of $\hat{\f}_{t+1}$ within a neighbourhood  $\mathcal{U}(\hat{\f}_{t})$ (see the formal definition below) of $\hat{\f}_{t}$. This consideration is well suited for the principal curves setting since if observation $x_{t}$ is close to $\hat{\f}_{t}$, one can expect that the polygonal line which well fits observations $x_{s}, s=1,\dots, t$ lies in a neighbourhood of $\hat{\f}_{t}$. In addition, if each polygonal line $\f$ is regarded as an action, we no longer assume that all actions are available at all times, and allow the set of available actions to vary at each time. This is a model known as "sleeping experts (or actions)" in prior work \citep{ACFS2003,KNS2008}. In this setting, defining the regret with respect to the best action in the whole set of actions in hindsight remains difficult since that action might sometimes be unavailable. Hence it is natural to define the regret with respect to the best ranking of all actions in the hindsight according to their losses or rewards, and at each round one chooses among the available actions by selecting the one which ranks the highest. \cite{KNS2008} introduced this notion of regret and studied both the full-information (best action) and partial-information (multi-armed bandit) settings with stochastic and adversarial rewards and adversarial action availability. They pointed out that the \textbf{EXP4} algorithm \citep{ACFS2003} attains the optimal regret in adversarial rewards case but has a runtime exponential in the number of all actions. \cite{KMB2009} considered full and partial information with stochastic action availability and proposed an algorithm that runs in polynomial time. In what follows, we  materialize our implementation by resorting to ''sleeping experts'' \emph{i.e.,} a special set of available actions that adapts to the setting of principal curves.
\medskip

Let $\sigma$ denote an ordering of $|\mathcal{F}_{p}|$ actions, and $\mathcal{A}_{t}$ a subset of the available actions at round $t$. We let $\sigma(\mathcal{A}_{t})$ denote the highest ranked action in $\mathcal{A}_{t}$. In addition, for any action $\f \in \mathcal{F}_{p}$ we define the reward $r_{\f,t}$ of $\f$ at round $t, t\geq 0$ by
\begin{equation*}\label{eq:def of reward}
    r_{\f, t} = c_{0} - \Delta(\f,x_{t}).
\end{equation*}
It is clear that $r_{\f, t} \in (0, c_{0})$. The convention from losses to gains is done in order to facilitate the subsequent performance analysis. The reward of an ordering $\sigma$ is the cumulative reward of the selected action at each time
\begin{equation*}
    \sum_{t=1}^{T}r_{\sigma(\mathcal{A}_{t}), t}, 
\end{equation*}
and the reward of the best ordering is $\max_{\sigma}\sum_{t=0}^{T}r_{\sigma(\mathcal{A}_{t}), t}$ (respectively, $\E\left[\max_{\sigma}\sum_{t=1}^{T}r_{\sigma(\mathcal{A}_{t}), t}\right]$ when $\At$ is stochastic).
\medskip

   
Our procedure starts with a \textbf{partition} step which aims at identifying the "relevant" neighbourhood of an observation $x\in\R^d$ with respect to a given polygonal line, and then proceeds with the definition of the \textbf{neighbourhood} of an action $\f$. We then provide the full implementation and prove a regret bound.

\begin{figure}[h]
    \centering
    \includegraphics[scale=0.25]{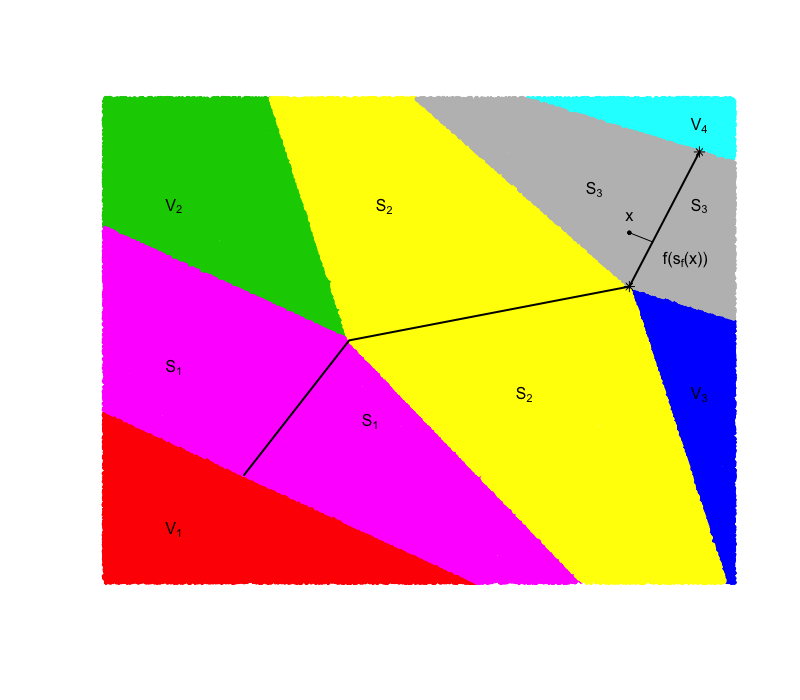}
    \caption{An example of a Voronoi partition.}
    \label{fig:voronoi_partition}
\end{figure}

\textbf{Partition} For any polygonal line $\f$ with $k$ segments, we denote by $\Vf = \left(v_{1}, \dots, v_{k+1}\right) $ its vertices and by $s_{i}, i=1, \dots, k$ the line segments connecting $v_{i}$ and $v_{i+1}$. In the sequel, we use $\f(\Vf)$ to represent the polygonal line formed by connecting consecutive vertices in $\Vf$ if no confusion arises. Let $V_{i}, i = 1,\dots, k+1$ and $S_{i}, i= 1,\dots, k$ be the Voronoi partitions of $\R^{d}$ with respect to $\f$, \emph{i.e.,} regions consisting of all points closer to vertex $v_{i}$ or segment $s_{i}$. \autoref{fig:voronoi_partition} shows an example of Voronoi partition with respect to $\f$ with 3 segments. 

\textbf{Neighbourhood} For any $x \in \R^{d}$, we define the neighbourhood $\mathcal{N}(x)$ with respect to $\f$ as the union of all Voronoi partitions whose closure intersects with two vertices connecting the projection $\f(s_{\f}(x))$ of $x$ to $\f$. For example, for the point $x$ in \autoref{fig:voronoi_partition}, its neighbourhood $\mathcal{N}(x)$ is the union of $S_{2}, V_{3}, S_{3}$ and $V_{4}$. In addition, let $\mathcal{N}_{t}(x) = \left\{x_{s} \in \mathcal{N}\left(x\right), s = 1,\dots, t. \right\}
$ be the set of observations $x_{1:t}$ belonging to $\mathcal{N}\left(x\right)$ and $\bar{\mathcal{N}}_{t}(x)$ be its average. Let $\mathcal{D}(M) = \sup_{x,y \in M}||x-y||_{2}$ denote the diameter of set $M \subset \R^d$. We finally define the local grid  $\mathcal{Q}_{\delta, t}(x)$ of $x \in \R^d$ at time $t$ as
\begin{equation*}
    \mathcal{Q}_{\delta,t}(x) = B\left(\bar{\mathcal{N}}_{t}(x),   \mathcal{D}\left(\mathcal{N}_{t}(x\right)\right) \cap \mathcal{Q}_{\delta}.
\end{equation*}
We can finally proceed to the definition of the neighbourhood $\mathcal{U}(\hat{\f}_{t})$ of $\hat{\f}_{t}$. Assume $\hat{\f}_{t}$ has $k_{t}+1$ vertices $\Vf = (\underbrace{v_{1:i_{t}-1}}_{(i)},
    \underbrace{v_{i_{t}:j_{t}-1}}_{(ii)}, 
    \underbrace{v_{j_{t}:k_{t}+1}}_{(iii)})$, where vertices of $(ii)$ belong to $\mathcal{Q}_{\delta,t}(x_{t})$ while those of $(i)$ and $(iii)$ do not. The neighbourhood $\mathcal{U}(\hat{\f}_{t})$ consists of $\f$ sharing vertices $(i), (iii)$ with  $\hat{\f}_{t}$, but can be equipped with different vertices $(ii)$ in $\mathcal{Q}_{\delta, t}(x_{t})$, \emph{i.e.}, 
    \begin{equation*}
    \mathcal{U}(\hat{\f}_{t}) = \left\{\f(\Vf), \quad \Vf = \left(v_{1:i_{t}-1}, v_{1:m}, v_{j_{t}:k_{t}+1}\right)\right\},
    \end{equation*}
    where $v_{1:m} \in \mathcal{Q}_{\delta,t}(x_{t})$ and $m$ is given by
    \begin{equation*}
    m = 
    \begin{cases}
   j_{t}-i_{t}-1 &\mbox{reduce segments by 1 unit},\\
   j_{t}-i_{t} &\mbox{same number of segments},\\
   j_{t}-i_{t}+1 &\mbox{increase segments by 1 unit}.
\end{cases}
\end{equation*}

\begin{algorithm}[ht]
\caption{A locally greedy algorithm to sequentially learn principal curves}
\label{proc:procedure 3}
\begin{algorithmic}[1]
\\\textbf{Input parameters}: $p >0$, $R >0$, $L > 0$, $\epsilon >0$, $\alpha >0$, $1>\beta >0$ and any penalty function $h$
\\\textbf{Initialization}: Given $(x_{t})_{1:t_{0}}$, obtain $\hat{\f}_{1}$ as the first principal component
\State \textbf{For $t=2,\dots,T$}
\State\hspace{\algorithmicindent} Draw $I_{t} \sim Bernoulli(\epsilon)$ and $z_{\f} \sim \pi$.
\State\hspace{\algorithmicindent} Let $$\hat{\sigma}_{t} = \text{sort}\left(\f, \quad \sum_{s=1}^{t-1}\hat{r}_{\f, s} -\frac{1}{\eta_{t-1}}h(\f) + \frac{1}{\eta_{t-1}}z_{\f}\right),$$ \emph{i.e.,} sorting all $\f \in \mathcal{F}_{p}$ in descending order according to their perturbed cumulative reward till $t-1$.
\State\hspace{\algorithmicindent} If $I_{t} = 1$, set $\mathcal{A}_{t} = \mathcal{F}_{p}$ and $\hat{\f}_{t} = \hat{\sigma}^{t}(\At)$ and observe $r_{\hat{\f}_{t}, t}$
\State\hspace{\algorithmicindent}\hspace{\algorithmicindent}\hspace{\algorithmicindent} 
\begin{equation*}
    \hat{r}_{\f,t} = r_{\f, t} \quad \text{for} \quad \f \in \mathcal{F}_{p}.
\end{equation*}
\State\hspace{\algorithmicindent} If $I_{t} = 0$, set $\mathcal{A}_{t} = \mathcal{U}(\hat{\f}_{t-1})$, $\hat{\f}_{t} = \hat{\sigma}^{t}(\At)$ and observe $r_{\hat{\f}_{t}, t}$
\State\hspace{\algorithmicindent}\hspace{\algorithmicindent}\hspace{\algorithmicindent} 
\begin{equation*}
    \hat{r}_{\f,t} = 
    \begin{cases}
    \frac{r_{\f,t}}{\mathbb{P}\left(\hat{\f}_{t} = \f|\mathcal{H}_{t}\right)} &\mbox{if \, $\f \in \mathcal{U}(\hat{\f}_{t-1}) \cap cond(t)$} \text{ and } \hat{\f}_{t} = \f, \\
   \alpha &\mbox{\text{otherwise}},
   \end{cases}
\end{equation*}
where $\mathcal{H}_{t}$ denotes all the randomness before time $t$ and $\mathit{cond}(t) = \left\{\f \in \mathcal{F}_{p}: \mathbb{P}\left(\hat{\f}_{t} = \f|\mathcal{H}_{t}\right) > \beta \right\}$. In particular, when $t=1$, we set $\hat{r}_{\f, 1} = r_{\f, 1}$ for all $\f \in \mathcal{F}_{p}$, $\mathcal{U}\left(\hat{\f}_{0}\right) = \emptyset$ and $\hat{r}_{\hat{\sigma}^{1}\left(\mathcal{U}\left(\hat{\f}_{0}\right)\right), 1} \equiv 0$.
\State \textbf{End for}
\end{algorithmic}
\end{algorithm}

In \autoref{proc:procedure 3}, we initiate the principal curve $\hat{\f}_{1}$ as the first component line segment whose vertices are the two farthest projections of data $x_{1:t_{0}}$ ($t_{0}$ can be set to 2 or 3 in practice) on the first component line. The reward of $\f$ at round $t$ in this setting is therefore $r_{\f,t} = c_{0} - \Delta(\f, x_{t_{0}+t})$. \autoref{proc:procedure 3} has an exploration phase (when $I_{t} = 1$) and an exploitation phase ($I_{t} = 0$). In the exploration phase,
it is allowed to observe rewards of all actions and 
 to choose an optimal perturbed action from the set $\mathcal{F}_{p}$ of all actions. In the exploitation phase, only rewards of a part of actions can be accessed and rewards of others are estimated by a constant, and we update our action from the neighbourhood $\mathcal{U}\left(\hat{\f}_{t-1}\right)$ of the previous action $\hat{\f}_{t-1}$. This local update (or search) greatly reduces computation complexity since $|\mathcal{U}(\hat{\f}_{t-1})| \ll \left|\mathcal{F}_{p}\right|$ when $p$ is large. In addition, this local search will be enough to account for the case when $x_{t}$ locates in $\mathcal{U}\left(\hat{\f}_{t-1}\right)$. The parameter $\beta$ needs to be carefully calibrated since it should not be too large to ensure that the condition $cond(t)$ is non-empty, otherwise all rewards are estimated by the same constant and thus lead to the same descending ordering of tuples for both $\left(\sum_{s=1}^{t-1}\hat{r}_{\f,s}, \f \in \mathcal{F}_{p}\right)$ and $\left(\sum_{s=1}^{t}\hat{r}_{\f,s}, \f \in \mathcal{F}_{p}\right)$. Therefore, we may face the risk of having $\hat{\f}_{t+1}$ in the neighbourhood of $\hat{\f}_{t}$ even if we are in the exploration phase at time $t+1$. Conversely, very small $\beta$ could result in large bias for the estimation $\frac{r_{\f,t}}{\mathbb{P}\left(\hat{\f}_{t} = \f|\mathcal{H}_{t}\right)}$ of $r_{\f,t}$. Note that the exploitation phase is close yet different to the label efficient prediction \citep[][Remark 1.1]{NGG2005} since we allow an action at time $t$ to be different from the previous one. \cite{NB2013} have proposed the \emph{Geometric Resampling} method to estimate the conditional probability $\mathbb{P}\left(\hat{\f}_{t} = \f|\mathcal{H}_{t}\right)$ since this quantity often does not have an explicit form. However, due to the simple exponential distribution of $z_{\f}$ chosen in our case, an explicit form of  $\mathbb{P}\left(\hat{\f}_{t} = \f|\mathcal{H}_{t}\right)$ is straightforward. 
\begin{theo}\label{thm:theorem 3}
Assume that $p >6$, $T \geq 2|\mathcal{F}_{p}|^2$ and let $\beta = \left|\mathcal{F}_{p}\right|^{-\frac{1}{2}}T^{-\frac{1}{4}}$, $\alpha = \frac{c_{0}}{\beta}$, $\hat{c}_{0} = \frac{2c_{0}}{\beta}$, $\epsilon = 1- \left|\mathcal{F}_{p}\right|^{\frac{1}{2} - \frac{3}{p}}T^{-\frac{1}{4}}$ and 
\begin{equation*}
    \eta_{1} = \eta_{2} =\dots=\eta_{T}= \frac{\sqrt{c_{1}p+c_{2}L+c_{3}}}{\sqrt{T(e-1)}\hat{c}_{0}}.
\end{equation*}
Then the procedure described in  \autoref{proc:procedure 3} satisfies the regret bound
\begin{equation*}
    \sum_{t=1}^{T}\E\left[\Delta\left(\hat{\f}_{t}, x_{t}\right)\right] \leq \inf_{\f \in \mathcal{F}_{p}}\E\left[\sum_{t=1}^{T}\Delta\left(\f, t\right)\right]  
    + \mathcal{O}(T^{\frac{3}{4}}).
\end{equation*}
\end{theo}
The proof of \autoref{thm:theorem 3} is presented in \autoref{sec:proofs}. The regret is upper bounded by a term of order $\left(\left|\mathcal{F}_{p}\right|^{\frac{1}{2}}T^{\frac{3}{4}}\right)$, sublinear in $T$. The term $(1-\epsilon)c_{0}T = c_{0}\left|\mathcal{F}_{p}\right|^{\frac{1}{2}}T^{\frac{3}{4}}$ is the price to pay for the local search (with a proportion $1-\epsilon$) of polygonal line $\hat{\f}_{t}$ in the neighbourhood of the previous $\hat{\f}_{t-1}$. If $\epsilon = 1$, we would have that  $\hat{c}_{0} = c_{0}$ and the last two terms in the first inequality of \autoref{thm:theorem 3} would vanish, hence the upper bound reduces to \autoref{thm:theorem 2}. In addition, our algorithm achieves an order that is smaller (from the perspective of both the number $\left|\mathcal{F}_{p}\right|$ of all actions and the total rounds $T$) than \cite{KMB2009} since at each time, the availability of actions for our algorithm can be either the whole action set or a neighbourhood of the previous action while \cite{KMB2009} consider at each time only partial and independent stochastic available set of actions generated from a predefined distribution.

\section{Numerical experiments}\label{sec:simus}

We illustrate the performance of \autoref{proc:procedure 3} on synthetic and real-life data. Our implementation (hereafter denoted by \texttt{slpc} -- Sequential Learning of Principal Curves) is conducted with the R language and thus our most natural competitor is the R package \texttt{princurve} \citep[which is the algorithm from][]{HS1989}. We let $p =20$, $R = \max_{t=1,\dots,T}||x||_{2}/\sqrt{d}$, $L = 0.01p \sqrt{d}R$. The spacing $\delta$ of the lattice is ajusted with respect to data scale.
\medskip

\textbf{Synthetic data} We generate a data set $\left\{x_{t}\in \R^2, t = 1, \dots,100\right\}$ uniformly along the curve $y = 0.05\times(x-5)^3$, $x\in [0,10]$. \autoref{tab:cumulative loss} shows the regret for the ground truth (sum of squared distances of all points to the true curve), \texttt{princurve} (sum of squared distances between observation $t+1$ and fitted \texttt{princurve} trained on all past $t$ observations) and \texttt{slpc} ($\sum_{t=0}^{T-1}\Delta(\hat{\f}_{t+1}, x_{t+1})$). \texttt{slpc} greatly outperforms \texttt{princurve} on this example, as illustrated by \autoref{fig:fig5} and \autoref{fig:fig6}. 
\begin{table}[h]
    \centering
    \begin{tabular}{ccc}
         \emph{ground truth} & \texttt{princurve} & \texttt{slpc}  \\ \hline
         0.945 (0) & 25.387 (0) &9.893 (0.246) \\  
    \end{tabular}
    \caption{Regret (cumulative loss) on synthetic data (average over 10 trials, with standard deviation in brackets). \texttt{princurve} is deterministic, hence the zero standard deviation.}
    \label{tab:cumulative loss}
\end{table}

\begin{figure}[ht]
\centering

\subfigure[\texttt{slpc}, $t=199$, 1st and 2nd coordinates]{\includegraphics[width=4cm]{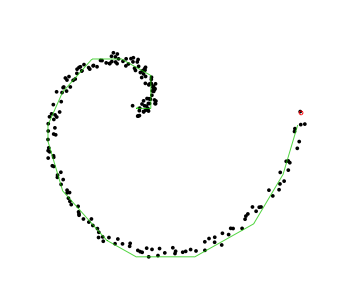}} \hfill
\subfigure[\texttt{slpc}, $t=199$, 3th and 5th coordinates]{\includegraphics[width=4cm]{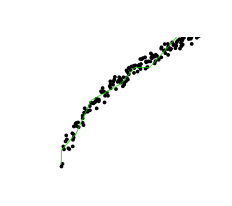}} \hfill
\subfigure[\texttt{slpc}, $t=199$, 4th and 6th coordinates]{\includegraphics[width=4cm]{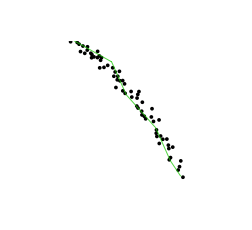}} 

\caption{\texttt{slpc} (green line) on synthetic data in higher dimension from different perspectives. Black dots represent recordings $x_{1:99}$, the red dot is the new recording $x_{200}$.}
\label{fig:fig9}
\end{figure}

\textbf{Synthetic data in high dimension} We also apply our algorithm on a data set $\left\{x_{t} \in \mathbb{R}^{6}, t = 1,2,\dots, 200\right\}$ in higher dimension. It is generated uniformly along a parametric curve whose coordinates are 
$$
\begin{pmatrix}
  0.5t\cos(t)   \\
  0.5t\sin(t)   \\
  0.5t     \\
  -t         \\
  \sqrt{t} \\
  2\ln (t+1)
\end{pmatrix}
$$
where $t$ takes 100 equidistant values in $[0, 2\pi]$. To the best of our knowledge, \cite{HS1989}, \cite{KEG1999} and \cite{BA2012} only tested their algorithm on 2-dimensional data. This example aims at illustrating that our algorithm also works on higher dimensional data. \autoref{tab:cumulative loss in higher dimension} shows the regret for the ground truth, \texttt{princurve} and \texttt{slpc}.
\begin{table}[h]
    \centering
    \begin{tabular}{ccc}
         \emph{ground truth} & \texttt{princurve} & \texttt{slpc}  \\ \hline
         3.290 (0) & 14.204 (0) & 6.797 (0.409) \\  
    \end{tabular}
    \caption{Regret (cumulative loss) on synthetic data in higher dimension (average over 10 trials, with standard deviation in brackets). \texttt{princurve} is deterministic.}
    \label{tab:cumulative loss in higher dimension}
\end{table}
In addition, \autoref{fig:fig9} shows the behaviour of \texttt{slpc} (green) on each dimension.

\begin{figure}[ht]
\centering
\subfigure[\texttt{princurve}, $t=100$]{\includegraphics[width=7cm]{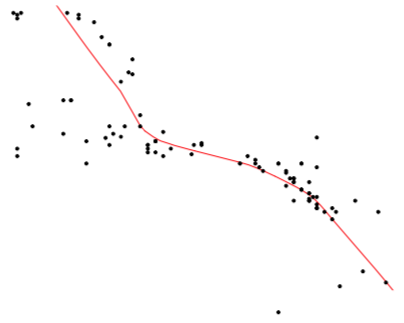}} \hfill
\subfigure[\texttt{princurve}, $t=125$]{\includegraphics[width=7cm]{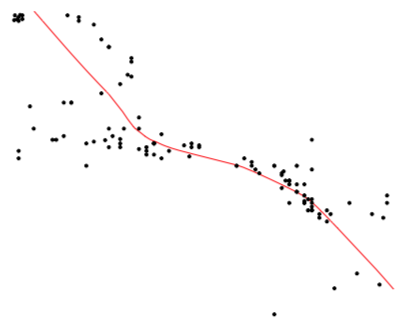}}
\subfigure[\texttt{slpc}, $t=100$]{\includegraphics[width=7cm]{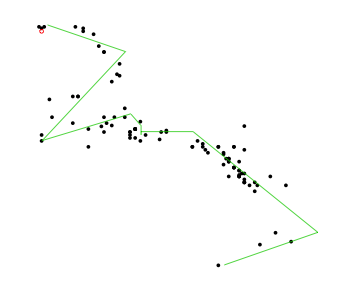}} \hfill
\subfigure[\texttt{slpc}, $t=125$]{\includegraphics[width=7cm]{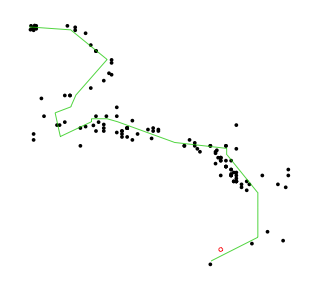}}
\caption{Seismic data. Black dots represent seismic recordings $x_{1:t}$, red dot is the new recording $x_{t+1}$.}
\label{fig:fig8}
\end{figure}

\begin{figure}[h]
\centering
\subfigure[$t=75$, \texttt{princurve}]{\includegraphics[width=7cm]{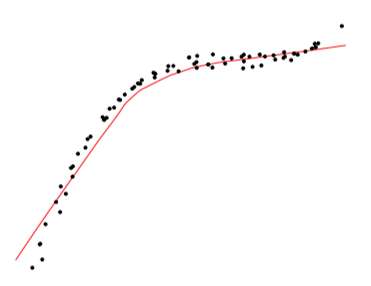}}
\hfill
\subfigure[$t=100$, \texttt{princurve}]{\includegraphics[width=7cm]{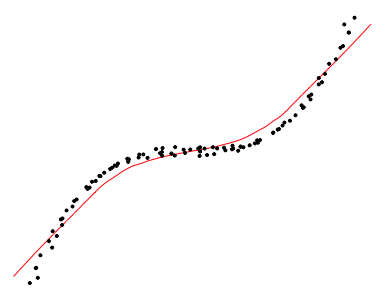}} \hfill
\subfigure[$t=75$, \texttt{slpc}]{\includegraphics[width=7cm]{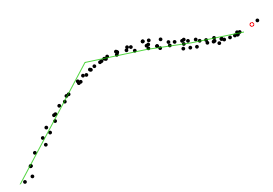}} \hfill
\subfigure[$t=100$, \texttt{slpc}]{\includegraphics[width=7cm]{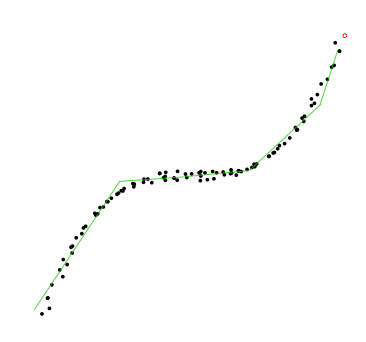}}
\caption{Synthetic data - Black dots represent data $x_{1:t}$, red point is the new observation $x_{t+1}$. \texttt{princurve} (solid red) and \texttt{slpc} (solid green).}
\label{fig:fig5}
\end{figure}

\begin{figure}[h]
\centering
\subfigure[At time $t=97$]{\includegraphics[width=7cm]{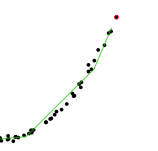}} \hfill
\subfigure[And at time $t=98$]{\includegraphics[width=7cm]{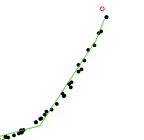}}
\caption{Synthetic data - Zooming in: how a new data point impacts only locally the principal curve.}
\label{fig:fig6}
\end{figure}

\textbf{Seismic data} Seismic data spanning long periods of time are essential for a thorough understanding of earthquakes. The “Centennial Earthquake Catalog” \citep{EV2002} aims at providing a realistic picture of the seismicity distribution on Earth. It consists in a global catalog of locations and magnitudes of instrumentally recorded earthquakes from 1900 to 2008. We focus on a particularly representative seismic active zone (a lithospheric border close to Australia) whose longitude is between E$130^\circ$ to E$180^\circ$ and latitude between S$70^\circ$ to N$30^\circ$, with $T=218$ seismic recordings. As shown in \autoref{fig:fig8}, \texttt{slpc} recovers nicely the tectonic plate boundary.
Lastly, since no ground truth is available, we use the $R^2$ coefficient to assess the performance (residuals are replaced by the squared distance between data points and their projections onto the principal curve). The average over 10 trials is 0.990.

\paragraph{Back to the synthetic data setting}

\autoref{fig:fig5} presents the predicted principal curve $\hat{\f}_{t+1}$ for both \texttt{princurve} (red) and \texttt{slpc} (green). The output of \texttt{princurve} yields a curve which does not pass in "the middle of data" but rather bends towards the curvature of the data cloud: \texttt{slpc} does not suffer from this behavior. To better illustrate the way \texttt{slpc} works between two epochs, \autoref{fig:fig6} focuses on the impact of collecting a new data point on the principal curve. We see that only a local vertex is impacted, whereas the rest of the principal curve remains unaltered. This cutdown in algorithmic complexity is one the key assets of \texttt{slpc}.

\begin{figure}[h]
\centering
\includegraphics[width=\textwidth]{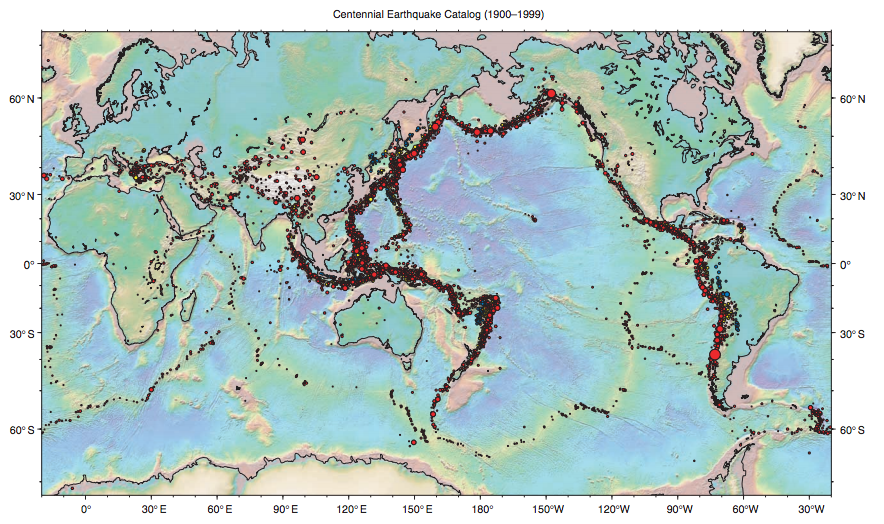}
\caption{Seismic data from \url{https://earthquake.usgs.gov/data/centennial/}}
\label{fig:fig7}
\end{figure}

\paragraph{Back to seismic data}
\autoref{fig:fig7} is taken from the USGS website\footnote{\href{https://earthquake.usgs.gov/data/centennial/}{https://earthquake.usgs.gov/data/centennial/}} and gives the global earthquakes locations on the period 1900--1999. The seismic data (latitude, longitude, magnitude of earthquakes, \emph{etc.}) used in the present paper may be downloaded from this website.

\begin{figure}[ht!]
\centering
\subfigure[$t=10$, \texttt{princurve}]{\includegraphics[width=6.5cm]{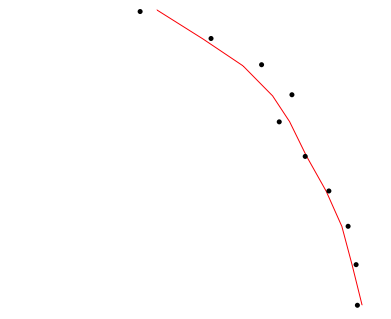}} \hfill
\subfigure[$t=127$, \texttt{princurve}]{\includegraphics[width=6.5cm]{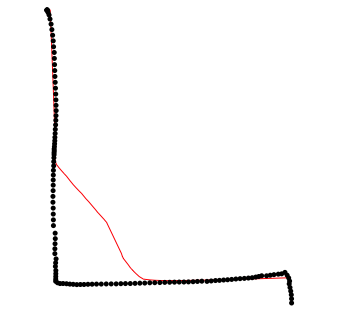}}
\subfigure[$t=10$, \texttt{slpc}]{\includegraphics[width=6.5cm]{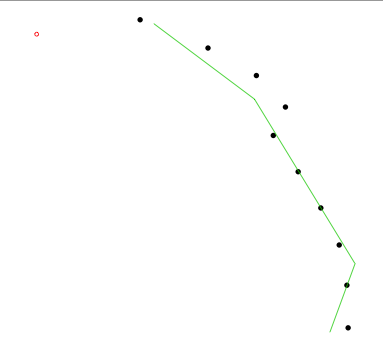}} \hfill
\subfigure[$t=127$, \texttt{slpc}]{\includegraphics[width=6.5cm]{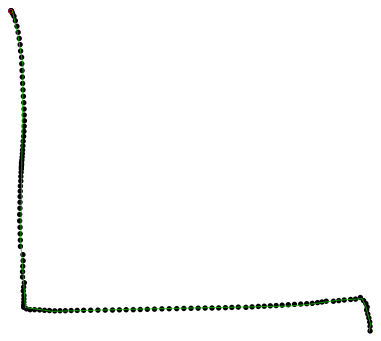}}
\caption{Daily commute data - Black dots represent collected locations $x_{1:t}$, red point is the new observation $x_{t+1}$. \texttt{princurve} (solid red) and \texttt{slpc} (solid green).}
\label{fig:commute data}
\end{figure}

\paragraph{Daily commute data} The identification of segments of personal daily commuting trajectories can help taxi or bus companies to optimise their fleets and increase frequencies on segments with high commuting activity. Sequential principal curves appear to be an ideal tool to address this learning problem: we test our algorithm on trajectory data from the University of Illinois at Chicago \footnote{\href{https://www.cs.uic.edu/~boxu/mp2p/gps_data.html}{https://www.cs.uic.edu/$\sim$boxu/mp2p/gps\_data.html}}. The data is obtained from the GPS reading systems carried by two of the lab members during their daily commute for 6 months in the Cook county and the Dupage county of Illinois. \autoref{fig:commute data} presents the learning curves yielded by \texttt{princurve} and \texttt{slpc} on geolocalization data for the first person, on May 30 in the data set. A particularly remarkable asset of \texttt{slpc} is that abrupt curvature in the data sequence is perfectly captured, whereas \texttt{princurve} does not enjoy the same flexibility. Again, we use the $R^2$ coefficient to assess the performance (where residuals are replaced by the squared distance between data points and their projections onto the principal curve). The average over 10 trials is 0.998.

\section{Proofs}\label{sec:proofs}
This section contains the proof of \autoref{thm:theorem 2} (note that \autoref{thm:theorem 1} is a straightforward consequence, with $\eta_{t} = \eta$, $t= 0,\dots, T$) and the proof of \autoref{thm:theorem 3} (which involves intermediary lemmas). Let us first define for each $t = 0, \dots, T$ the following forecaster sequence $(\hat{\f}^{\star}_{t})_t$
\begin{align*}
    & \hat{\f}_{0}^{\star} = \underset{\f \in \mathcal{F}_{p}}{\arg\inf}\, \left\{\Delta_{\f,0}\right\}= \underset{\f \in \mathcal{F}_{p}}{\arg\inf}\left\{ \frac{1}{\eta_{0}}h(\f) - \frac{1}{\eta_{0}}z_{\f}\right\},\\
    &\hat{\f}^{\star}_{t} = \underset{\f \in \mathcal{F}_{p}}{\arg\inf}\, \left\{\sum_{s=0}^{t}\Delta_{\f,s}\right\} = \underset{\f \in \mathcal{F}_{p}}{\arg\inf}\left\{\sum_{s=1}^{t}\Delta(\f, x_{s}) + \frac{1}{\eta_{t-1}}h(\f) - \frac{1}{\eta_{t-1}}z_{\f}\right\}, \quad t\geq 1.
\end{align*}
Note that $\hat{\f}^{\star}_{t}$ is an "illegal" forecaster since it peeks into the future. In addition, denote by 
\begin{equation*}
    \f^{\star} =  \underset{\f \in \mathcal{F}_{p}}{\arg\inf}\left\{\sum_{t=1}^{T}\Delta(\f, x_t) +  \frac{1}{\eta_{T}}h(\f)\right\}
\end{equation*}
the polygonal line in $\mathcal{F}_{p}$ which minimizes the cumulative loss in the first $T$ rounds plus a penalty term. $\f^{\star}$ is deterministic while $\hat{\f}^{\star}_{t}$ is a random quantity (since it depends on $z_{\f}$, $\f \in \mathcal{F}_{p}$ drawn from $\pi$). If several $\f$ attain the infimum, we choose $\f^{\star}_T$ as the one having the smallest complexity. We now enunciate the first (out of three) intermediary technical result.
\begin{lem}\label{lem:lemma 1}
For any sequence $x_{1}, \dots, x_T$ in $\cball$,
\begin{equation}\label{eq:induction}
\sum_{t=0}^{T}\Delta_{\hat{\f}_{t}^{\star}, t} \leq \sum_{t=0}^{T}\Delta_{\hat{\f}_{T}^{\star}, t},  \qquad \pi\text{-almost surely}.
\end{equation}
\end{lem}
\begin{proof}
Proof by induction on $T$. Clearly \eqref{eq:induction} holds for $T=0$. Assume that \eqref{eq:induction} holds for $T-1$:
\begin{equation*}
   \sum_{t=0}^{T-1}\Delta_{\hat{\f}_{t}^{\star}, t} \leq \sum_{t=0}^{T-1}\Delta_{\hat{\f}_{T-1}^{\star}, t}.    
\end{equation*}
Adding $\Delta_{\hat{\f}_{T}^{\star}, T}$ to both sides of the above inequality concludes the proof.
\end{proof}

By \eqref{eq:induction} and the definition of $\hat{\f}_{T}^{\star}$, for $k\geq 1$, we have  $\pi$-almost surely that
\begin{align*}
    \sum_{t=1}^{T}\Delta(\hat{\f}_{t}^{\star}, x_{t}) &\leq \sum_{t=1}^{T}\Delta(\hat{\f}_{T}^{\star}, x_{t}) +  \frac{1}{\eta_{T}}h(\hat{\f}_{T}^{\star}) - \frac{1}{\eta_{T}}Z_{\hat{\f}_{T}^{\star}} + \sum_{t=0}^{T}\left(\frac{1}{\eta_{t-1}}-\frac{1}{\eta_{t}}\right)\left(h(\hat{\f}_{t}^{\star})-Z_{\hat{\f}_{t}^{\star}}\right)\notag \\
    & \leq \sum_{t=1}^{T}\Delta(\f^{\star}, x_{t}) +  \frac{1}{\eta_{T}}h(\f^{\star}) - \frac{1}{\eta_{T}}Z_{\f^{\star}} + \sum_{t=0}^{T}\left(\frac{1}{\eta_{t-1}}-\frac{1}{\eta_{t}}\right)\left(h(\hat{\f}_{t}^{\star})-Z_{\hat{\f}_{t}^{\star}}\right) \notag \\  
    &= \inf_{\f \in \mathcal{F}_{p}}\left\{\sum_{t=1}^{T}\Delta(\f, x_{t}) +  \frac{1}{\eta_{T}}h(\f)\right\} - \frac{1}{\eta_{T}}Z_{\f^{\star}} +  \sum_{t=0}^{T}\left(\frac{1}{\eta_{t-1}}-\frac{1}{\eta_{t}}\right)\left(h(\hat{\f}_{t}^{\star})-Z_{\hat{\f}_{t}^{\star}}\right), 
\end{align*}
where $1/\eta_{-1} = 0$ by convention.
The second and third inequality is due to respectively the definition of $\hat{\f}_{T}^{\star}$ and $\f_{T}^{\star}$.
Hence
\begin{align*}
\mathbb{E}\left[\sum_{t=1}^{T}\Delta\left(\hat{\f}_{t}^{\star}, x_{t}\right)\right] &\leq \inf_{\f \in \mathcal{F}_{p}}\left\{\sum_{t=1}^{T}\Delta(\f, x_{t}) +  \frac{1}{\eta_{T}}h(\f)\right\} - \frac{1}{\eta_{T}}\E[Z_{\f^{\star}_{T}}] +  \sum_{t=0}^{T}\E\left[\left(\frac{1}{\eta_{t}}-\frac{1}{\eta_{t-1}}\right)\left(-h(\hat{\f}_{t}^{\star}) +Z_{\hat{\f}_{t}^{\star}}\right)\right] \\
& \leq \inf_{\f \in \mathcal{F}_{p}}\left\{\sum_{t=1}^{T}\Delta(\f, x_{t}) +  \frac{1}{\eta_{T}}h(\f)\right\} + \sum_{t=1}^{T}\left(\frac{1}{\eta_{t}}-\frac{1}{\eta_{t-1}}\right)\mathbb{E}\left[\sup_{\f \in \mathcal{F}_{p}}\left(-h(\f)+Z_{\f}\right)\right] \\
& = \inf_{\f \in \mathcal{F}_{p}}\left\{\sum_{t=1}^{T}\Delta(\f, x_{t}) +  \frac{1}{\eta_{T}}h(\f)\right\} + \frac{1}{\eta_{T}}\mathbb{E}\left[\sup_{\f \in \mathcal{F}_{p}}\left(-h(\f) +Z_{\f}\right)\right],
\end{align*}
where the second inequality is due to $\E[Z_{\f^{\star}_{T}}] = 0$ and $\left(\frac{1}{\eta_{t}}-\frac{1}{\eta_{t-1}}\right) >0$ for $t= 0, 1, \dots, T$ since $\eta_{t}$ is decreasing in $t$ in \autoref{thm:theorem 2}. In addition, for $y\geq 0$, one has
\begin{equation*}
\mathbb{P}\left(-h(\f) + Z_{\f} > y \right) = \e^{-h(\f)-y}.
\end{equation*}
Hence, for any $y \geq 0$

\begin{equation*}
\mathbb{P}\left(\sup_{\f \in \mathcal{F}_{p}}\left(-h(\f)+ Z_{\f}\right) > y\right) \leq \sum_{\f \in \mathcal{F}_{p}} \mathbb{P}\left(Z_{\f} \geq h(\f) + y\right) = \sum_{\f \in \mathcal{F}_{p}} \e^{-h(\f)}\e^{-y} = u \e^{-y},
\end{equation*}
where $u = \sum_{\f \in \mathcal{F}_{p}}\e^{-h(\f)}$. Therefore, we have
\begin{align*}
   \E\left[\sup_{\f \in \mathcal{F}_{p}}\left(-h(\f)+Z_{\f}\right) - \ln u\right] &\leq \E\left[ \max \left(0, \sup_{\f \in \mathcal{F}_{p}}\left(-h(\f)+Z_{\f}-\ln u\right) \right)\right]\\
    & \leq \int_{0}^{\infty}\mathbb{P}\left( \max \left(0, \sup_{\f \in \mathcal{F}_{p}}\left(-h(\f)+Z_{\f}-\ln u\right) \right) > y\right)\mathrm{d}y \\
    & \leq \int_{0}^{\infty}\mathbb{P}\left(\sup_{\f \in \mathcal{F}_{p}}\left(-h(\f)+Z_{\f}\right) > y + \ln u\right)\mathrm{d}y \\
    & \leq \int_{0}^{\infty} u \e^{-(y+\ln u)}\mathrm{d}y = 1.\\
\end{align*}
We thus obtain
\begin{equation}\label{eq:lem 1}
    \mathbb{E}\left[\sum_{t=1}^{T}\Delta\left(\hat{\f}_{t}^{\star}, x_{t}\right)\right]\leq \inf_{\f \in \mathcal{F}_{p}}\left\{\sum_{t=1}^{T}\Delta(\f, x_{t}) +  \frac{1}{\eta_{T}}h(\f)\right\} + \frac{1}{\eta_{T}}\left(1 +\ln \sum_{\f \in \mathcal{F}_{p}}\e^{-h(\f)}\right).
\end{equation}
Next, we control the regret of \autoref{proc:procedure 2}.
\begin{lem}\label{lem:lemma 2}
Assume that $z_{\f}$ is sampled from the symmetric exponential distribution in $\R$, \emph{i.e.}, $\pi(z) = \e^{-z}\mathbbm{1}_{\{z >0 \}}$. Assume that $\sup_{t=1,\dots, T}\eta_{t-1}\leq \frac{1}{d(2R+\delta)^2}$, and define $c_{0}= d(2R+\delta)^2$. Then for any sequence $(x_{t}) \in \cball$, $t=1,\dots, T$,
\begin{equation}\label{eq:lem 2}
    \sum_{t=1}^{T}\E\left[\Delta\left(\hat{\f}_{t}, x_{t}\right)\right] \leq \sum_{t=1}^{T}\left(1+\eta_{t-1}c_{0}(\e-1)\right)\E\left[\Delta\left(\hat{\f}_{t}^{\star}, x_{t}\right)\right].
\end{equation}
\end{lem}

\begin{proof}
Let us denote by 
\begin{equation*}
    F_{t}(Z_{\f}) = \Delta\left(\hat{\f}_{t}, x_{t}\right) = \Delta\left(\farginf \left(\sum_{s=1}^{t-1}\Delta(\f, x_{s}) +  \frac{1}{\eta_{t-1}}h(\f) - \frac{1}{\eta_{t-1}}
    Z_{\f}
    \right),  x_{t}\right)
\end{equation*} 
the instantaneous loss suffered by the polygonal line $\hat{\f}_{t}$ when $x_{t}$ is obtained. We have
\begin{align*}
    \E[\Delta\left(\hat{\f}_{t}^{\star}, x_{t}\right)] &= \int F_{t}\left(z -\eta_{t-1} \Delta\left(\f, x_{t}\right)\right)\pi(z)\mathrm{d}z \\ &= \int F_{t}(z)\pi\left(z+ \eta_{t-1}\Delta(\f, x_{t})\right)\mathrm{d}z \\
    & = \int F_{t}(z)\e^{-\left(z+\eta_{t-1}\Delta(\f, x_{t})\right)}\mathrm{d}z \\
    &\geq \e^{-\eta_{t-1} d(2R+\delta)^2}\int F_{t}(z)\e^{- z}\mathrm{d}z \\ &= \e^{-\eta_{t-1} d(2R+\delta)^2}\E[\Delta\left(\hat{\f}_{t}, x_{t}\right)],
\end{align*}
where the inequality is due to the fact that $\Delta(\f, x) \leq d(2R+\delta)^2$ holds uniformly for any $\f \in \mathcal{F}_{p}$ and $x \in \cball$.
Finally, summing on $t$ on both sides and using the elementary inequality $\e^{x} \leq 1+(\e-1)x$ if $x \in (0,1)$ concludes the proof.
\end{proof}

\begin{lem}\label{lem:lemma 3}
For $k \in \llbracket1, p\rrbracket$, we control the cardinality of set $\big\{\f \in \mathcal{F}_{p}, \K(\f) = k\big\}$ as
\begin{align*}
    \ln \left|\big\{\f \in \mathcal{F}_{p}, \mathcal{K}(\f) = k\big\}\right| &\leq   \left(\ln(8p\e V_{d}) + 3d^{\frac{3}{2}}-d\right)k + \left(\frac{\ln 2}{\delta\sqrt{d}} + \frac{d}{\delta}\right)L +  d\ln\left(\frac{\sqrt{d}(2R+\delta)}{\delta}\right)\\
    &\overset{\Delta}{=}c_{1}k+c_{2}L + c_{3},
\end{align*}
where $V_{d}$ denotes the volume of the unit ball in $\R^{d}$.
\end{lem}
\begin{proof}
 First, let $N_{k, \delta}$ denote the set of polygonal lines with $k$ segments and whose vertices are in $\mathcal{Q}_{\delta}$. Notice that $N_{k, \delta}$ is different from $\{\f \in \mathcal{F}_{p}, \K(\f) = k\}$ and that $$\left|\{\f \in \mathcal{F}_{p}, \K(\f) = k\}\right|\leq \binom{p}{k}\left|N_{k, \delta}\right|.$$
 Hence
 \begin{align*}
     \ln \left|\{\f \in \mathcal{F}_{p}, \K(\f) = k\}\right| &\leq \ln \binom{p}{k} + \ln \left|N_{k, \delta}\right| \\
     & \leq k\ln \frac{p\e}{k} + k \left(\ln 8V_{d} + 3d^{\frac{3}{2}}-d\right) + \left(\frac{\ln 2}{\sqrt{d}\delta} + \frac{d}{\delta}\right)L +  d\ln\left(\frac{\sqrt{d}(2R+\delta)}{\delta}\right)\\
     &\leq k\ln (p\e) + k \left(\ln 8V_{d} + 3d^{\frac{3}{2}}-d\right) + \left(\frac{\ln 2}{\sqrt{d}\delta} + \frac{d}{\delta}\right)L + d\ln\left(\frac{\sqrt{d}(2R+\delta)}{\delta}\right),
 \end{align*}
where the second inequality is a consequence to the elementary inequality $\binom{p}{k} \leq 
\left(\frac{p\e}{k}\right)^{k}$ combined with Lemma 2 in \cite{KEG1999}.
\end{proof}

We now have all the ingredients to prove  \autoref{thm:theorem 1} and \autoref{thm:theorem 2}.
\medskip

 First, combining \eqref{eq:lem 1} and \eqref{eq:lem 2} yields that
\begin{align*}\label{eq:temp 1}
    \sum_{t=1}^{T}\E\left[\Delta(\hat{\f}_{t}, x_{t})\right] &\leq \inf_{\f \in \mathcal{F}_{p}}\left\{\sum_{t=1}^{T}\Delta(\f, x_{t}) +  \frac{1}{\eta_{T}}h(\f)\right\} + \frac{1}{\eta_{T}}\left(\frac{1}{2}+\ln \sum_{\f \in \mathcal{F}_{p}}\e^{-h(\f)}\right) \\  &\qquad\qquad + c_{0}(\e-1)\sum_{t=1}^{T}\eta_{t-1}\E\left[\Delta(\hat{\f}_{t}^{\star}, x_{t})\right] \notag \\
    &\leq \inf_{k \in \llbracket1,p\rrbracket}\left\{\inf_{\substack{\f \in \mathcal{F}_{p} \\ \mathcal{K}(\f) = k}}\left\{\sum_{t=1}^{T}\Delta(\f, x_{t}) + \frac{h(\f)}{\eta_{T}}\right\}\right\} + \frac{1}{\eta_{T}}\left(\frac{1}{2}+\ln \sum_{\f \in \mathcal{F}_{p}}\e^{-h(\f)}\right) \\ &\qquad\qquad + c_{0}(\e-1)\sum_{t=1}^{T}\eta_{t-1}\E\left[\Delta(\hat{\f}_{t}^{\star}, x_{t})\right].
    \end{align*}
Assume that $\eta_{t} = \eta$, $t=0,\dots,T$ and $h(\f) = c_{1}\mathcal{K}(\f)+c_{2}L+c_{3}$ for $\f \in \mathcal{F}_{p}$, then $\left(\frac{1}{2} + \sum_{\f \in \mathcal{F}_{p}} \e^{-h(\f)}\right) \leq 0$ and moreover
\begin{align*}
    \sum_{t=1}^{T}\E\left[\Delta(\hat{\f}_{t}, x_{t})\right] &\leq S_{T, h, \eta} + \frac{1}{\eta}\left(\frac{1}{2}+\ln \sum_{\f \in \mathcal{F}_{p}}\e^{-h(\f)}\right) + c_{0}(\e-1)\eta\sum_{t=1}^{T}\E\left[\Delta(\hat{\f}_{t}^{\star}, x_{t})\right] \\
    & \leq S_{T, h, \eta} + c_{0}(\e-1)\eta S_{T, h, \eta}  \\
    & \leq S_{T, h, \eta} + \eta c_{0}(\e-1) \inf_{\f \in \mathcal{F}_{p}} \sum_{t=1}^{T}\Delta(\f,x_{t}) +c_{0}(\e-1)(c_{	1}p + c_{2}L+ c_{3}),
\end{align*}
where $$S_{T, h, \eta} = \inf_{k \in \llbracket1,p\rrbracket}\left\{\inf_{\substack{\f \in \mathcal{F}_{p} \\ \mathcal{K}(\f) = k}}\left\{\sum_{t=1}^{T}\Delta(\f, x_{t}) + \frac{h(\f)}{\eta}\right\}\right\}$$ and the second inequality is obtained with \autoref{lem:lemma 1}.
By setting $$\eta = \sqrt{\frac{c_{1}p + c_{2}L+ c_{3}}{c_{0}(\e-1)\inf_{\f \in \mathcal{F}_{p}} \sum_{t=1}^{T}\Delta(\f,x_{t})}}$$ we obtain
\begin{multline*}
    \sum_{t=1}^{T}\E\left[\Delta(\hat{\f}_{t}, x_{t})\right] \\ \leq \inf_{k \in \llbracket1,p\rrbracket}\left\{\inf_{\substack{\f \in \mathcal{F}_{p} \\ \mathcal{K}(\f) = k}}\left\{\sum_{t=1}^{T}\Delta(\f, x_{t}) + \sqrt{c_{0}(\e-1)r_{T,k,L}}\right\}\right\} 
    + \sqrt{c_{0}(\e-1)L_{T,p,L}} + c_{0}(\e-1)c_{1}p + c_{2}L + c_{3},
\end{multline*}
where $r_{T,k,L} = \inf_{\f \in \mathcal{F}_{p}} \sum_{t=1}^{T}\Delta(\f,x_{t})(c_{1}k + c_{2}L + c_{3})$. This proves \autoref{thm:theorem 1}.
\medskip

Finally, assume that $$\eta_{0} = \frac{\sqrt{c_{1}p+c_{2}L+c_{3}}}{c_{0}\sqrt{(\e-1)}}\quad \text{and}\quad \eta_{t} = \frac{\sqrt{c_{1}p+c_{2}L+c_{3}}}{c_{0}\sqrt{(\e-1)t}},\qquad t = 1, \dots, T.$$
Since $\E\left[\Delta(\hat{\f}_{t}^{\star}, x_{t})\right] \leq c_{0}$ for any $t = 1,\dots, T$, we have
\begin{align*}
    \sum_{t=1}^{T}\E\left[\Delta(\hat{\f}_{t}, x_{t})\right]& \leq \inf_{k \in \llbracket1,p\rrbracket}\left\{\inf_{\substack{\f \in \mathcal{F}_{p} \\ \mathcal{K}(\f) = k}}\left\{\sum_{t=1}^{T}\Delta(\f, x_{t}) + \frac{h(\f)}{\eta_{T}}\right\}\right\} + \frac{1}{\eta_{T}}\left(1+\ln \sum_{\f \in \mathcal{F}_{p}}\e^{-h(\f)}\right) \\ &\quad + c_{0}^2(\e-1)\sum_{t=1}^{T}\eta_{t-1}\\
    &\leq \inf_{k \in \llbracket1,p\rrbracket}\left\{\inf_{\substack{\f \in \mathcal{F}_{p} \\ \mathcal{K}(\f) = k}}\left\{\sum_{t=1}^{T}\Delta(\f, x_{t}) + c_{0}\sqrt{(\e-1)T(c_{0}k+c_2L+c_{3})}\right\}\right\} \\ & \qquad\qquad\qquad\qquad\qquad\qquad\qquad\qquad\qquad + 2c_{0}\sqrt{(\e-1)T(c_{0}p+c_2L+c_{3})},
\end{align*}
which concludes the proof of \autoref{thm:theorem 2}.

\begin{lem}\label{lem:lemma 4}
Using \autoref{proc:procedure 3},  if $0<  \epsilon \leq 1$, $ 0<\beta <1$, $\alpha \geq \frac{(1-\beta)c_{0}}{\beta}$ and  $\left|\mathcal{U}\left(\hat{\f}_{t-1}\right)\right| \geq 2$ for all $t \geq 2$, where $\left|\mathcal{U}\left(\hat{\f}_{t-1}\right)\right|$ is the cardinality of $\mathcal{U}\left(\hat{\f}_{t-1}\right)$, then we have

\begin{equation*}
 \cumrwrd \geq  \sum_{t=1}^{T}\exphatordonwhole - 2(1-\epsilon)\alpha \beta\sum_{t=1}^{T}\left|\mathcal{U}\left(\hat{\f}_{t-1}\right)\right|.  
\end{equation*}
\end{lem}
\begin{proof} First notice that $\mathcal{A}_{t} = \mathcal{U}\left(\hat{\f}_{t-1}\right)$ if $I_{t} = 0$, and that for $t \geq 2$
\begin{align*}
    \E \left[r_{\hat{\f}_{t},t} \bigg| \mathcal{H}_{t}, I_{t} = 0\right] =& \E\left[r_{\hat{\sigma}^{t} \left(\mathcal{A}_{t}\right), t}\bigg| \mathcal{H}_{t}, I_{t} =0\right] \notag \\
    = & \sum_{\f \in \mathcal{A}_{t} \cap {cond(t)}} r_{\f, t}\P \left(\hat{\sigma}^{t}\left(\mathcal{A}_{t} \right) = \f\, \bigg| \mathcal{H}_{t} \right) + \sum_{\f \in \mathcal{A}_{t} \cap {cond(t)}^{c}} r_{\f, t}\P \left(\hat{\sigma}^{t}\left(\mathcal{A}_{t} \right) = \f \, \bigg| \mathcal{H}_{t} \right) \notag \\
     \geq & \sum_{\f \in \mathcal{A}_{t} \cap {cond(t)}} r_{\f, t} + \sum_{\f \in \mathcal{A}_{t} \cap {cond(t)}^{c}}\alpha \P \left(\hat{\sigma}^{t}\left(\mathcal{A}_{t} \right) = \f \, \bigg| \mathcal{H}_{t} \right) \notag \\  & -  (1-\beta) \sum_{\f \in \mathcal{A}_{t} \cap {cond(t)}} r_{\f, t} \notag  - \sum_{\f \in \mathcal{A}_{t} \cap {cond(t)}^{c}}\left( \alpha - r_{\f,t} \right)\P \left(\hat{\sigma}^{t}\left(\mathcal{A}_{t} \right) = \f \, \bigg| \mathcal{H}_{t} \right) \notag \\
      = & \E \left[\hat{r}_{\hat{\sigma}^{t} \left(\mathcal{A}_{t}\right), t}\bigg| \mathcal{H}_{t}, I_{t} = 0\right] - (1-\beta) \sum_{\f \in \mathcal{A}_{t} \cap {cond(t)}} r_{\f, t} \notag  \\ & - \sum_{\f \in \mathcal{A}_{t} \cap {cond(t)}^{c}}\left( \alpha - r_{\f,t} \right)\P \left(\hat{\sigma}^{t}\left(\mathcal{A}_{t} \right) = \f \, \bigg| \mathcal{H}_{t} \right) \notag \\
      \geq & \E \left[\hat{r}_{\hat{\sigma}^{t} \left(\mathcal{A}_{t}\right), t}\bigg| \mathcal{H}_{t}, I_{t} = 0\right] - (1-\beta)c_{0}\left|\mathcal{A}_{t}\right| - \alpha \beta \left|\mathcal{A}_{t}\right| \notag \\
      \geq & \E \left[\hat{r}_{\hat{\sigma}^{t} \left(\mathcal{A}_{t}\right), t}\bigg| \mathcal{H}_{t}, I_{t} = 0\right] - 2\alpha \beta \left|\mathcal{A}_{t}\right|,
\end{align*}
where $cond(t)^{c}$ denotes the complement of set $cond(t)$. The first inequality above is due to the assumption that for all $\f \in \mathcal{A}_{t}\cap cond(t)$, we have $\P \left(\hat{\sigma}^{t}\left(\mathcal{A}_{t} \right) = \f \, \bigg| \mathcal{H}_{t} \right) \geq \beta$. For $t=1$, the above inequality is trivial since $\hat{r}_{\hat{\sigma}^{1}\left(\mathcal{U}\left(\hat{\f}_{0}\right)\right), 1} \equiv 0$ by its definition. Hence, for $t\geq 1$, one has
\begin{align}\label{ieq:ieq1 lemma 4}
   \condilocrwrd &= \epsilon   \mathbb{E}\left[\ordonwhole \bigg|\mathcal{H}_{t}, I_{t} = 1\right] + (1-\epsilon) \E\left[r_{\hat{\sigma}^{t} \left(\mathcal{A}_{t}\right), t}\bigg| \mathcal{H}_{t}, I_{t} =0\right] \notag \\ 
   &\geq \E \left[\hat{r}_{\hat{\f}_{t}, t}\bigg| \mathcal{H}_{t} \right] - 2\alpha \beta \left|\mathcal{A}_{t}\right|.
\end{align}
Summing on both sides of inequality \eqref{ieq:ieq1 lemma 4} over $t$ terminates the proof of \autoref{lem:lemma 4}.
\end{proof}
\medskip

\begin{lem}\label{lem:lemma 5}
Let $\hat{c}_{0}= \frac{c_{0}}{\beta} + \alpha$. If $ 0< \eta_{1} = \eta_{2} = \dots = \eta_{T} = \eta < \frac{1}{{\hat{c}}_{0}}$, then we have 
\begin{equation*}
    \expglbhatordonwholepenalty - \sum_{t=1}^{T}\exphatordonwhole \leq {\hat{c}}_{0}^{2}(e-1)\eta T + {\hat{c}}_{0}(e-1)\left(c_{1}p+c_{2}L+c_{3}\right).
\end{equation*}
\end{lem}
\begin{proof}
By the definition of $\hat{r}_{\f, t}$ in \autoref{proc:procedure 3}, for any $\f \in \mathcal{F}_{p}$ and $t \geq 1$, we have 
\begin{equation*}
    \hat{r}_{\f, t} \leq \max \left\{\frac{r_{\f, t}}{\P \left( \hat{\f}_{t} = \f \bigg| \mathcal{H}_{t} \right)}, \alpha, r_{\f, t} \right\} \leq \max \left\{\frac{c_{0}}{\beta}, \alpha \right\} \leq \hat{c}_{0},
\end{equation*}
where in the second inequality we use that $r_{\f,t} \leq c_{0}$ for all $\f$ and $t$, and that $\P \left( \hat{\f}_{t} = \f \bigg| \mathcal{H}_{t} \right) \geq \beta$ when $\f \in \mathcal{U}\left(\hat{\f}_{t-1}\right) \cap cond(t)$. The rest of the proof is similar to those of \autoref{lem:lemma 1} and \autoref{lem:lemma 2}. In fact, if we define by $\hat{\Delta}\left(\f, x_{t}\right) = \hat{c}_{0} - \hat{r}_{\f,t}$, then one can easily observe the following relation when $I_{t} = 1$ (similar relation in the case that $I_{t}$ = 0)

\begin{align*}
    \hat{\f}_{t} = \hat{\sigma}^{t}\left(\mathcal{F}_{p}\right) &= \argmax_{\f \in \mathcal{F}_{p}}\left\{\sum_{s=1}^{t-1}\hat{r}_{\f,s} + \frac{1}{\eta}\left(z_{\f}-h(\f)\right)\right\} \\
    & = \argmin_{\f \in \mathcal{F}_{p}}\left\{\sum_{s=1}^{t-1}\hat{\Delta}(\f, x_{s}) + \frac{1}{\eta}\left(h(\f) -z_{\f}\right)\right\}.
\end{align*}
Then applying \autoref{lem:lemma 1} and \autoref{lem:lemma 2} on this newly defined sequence $\hat{\Delta}\left(\hat{\f}_{t}, x_{t}\right), t=1,\dots T$ leads to the result of \autoref{lem:lemma 5}.

\end{proof}

The proof of the upcoming \autoref{lem:lemma 6} requires the following submartingale inequality: let $Y_{0}, \dots Y_{T}$ be a sequence of random variable adapted to random events $\mathcal{H}_{0}, \dots, \mathcal{H}_{T}$ such that for $1\leq t \leq T$, the following three conditions hold
\begin{equation*}
    \E\left[Y_{t}|H_{t}\right] \leq 0,\quad \mathrm{Var}(Y_{t}|H_{t}) \leq a^2, \quad Y_{t} - \E\left[Y_{t}|H_{t}\right] \leq b.
\end{equation*}
Then for any $\lambda >0$, 
\begin{equation*}
    \mathbb{P}\left(\sum_{t=1}^{T}Y_{t} > Y_{0} + \lambda \right) \leq \exp\left(-\frac{\lambda^2}{2T(a^2+b^2)}\right).
\end{equation*}
The proof can be found in \citet[][Theorem 7.3]{CL2006}.

\begin{lem}\label{lem:lemma 6}
Assume that $0 < \beta < \frac{1}{\left|\mathcal{F}_{p}\right|}, \alpha \geq \frac{c_{0}}{\beta}$ and $\eta >0$, then we have
\begin{multline*}
    \expglbordonwholepenalty - \expglbhatordonwholepenalty \\ \leq \left(1-\left|\mathcal{F}_{p}\right|\beta\right)\sqrt{2T\left[\frac{c_{0}^2}{\beta}+\alpha^2(1-\beta) +\left(c_{0}+2\alpha\right)^2\right]\ln \left(\frac{1}{\beta}\right)} + \left|\mathcal{F}_{p}\right|\beta c_{0}T.
\end{multline*}
\end{lem}

\begin{proof}
First, we have almost surely that
\begin{align*}
    \glbordonwholepenalty - \glbhatordonwholepenalty &\leq \max_{\f \in \mathcal{F}_{p}}\sum_{t=1}^{T}\left(r_{\f, t} -\hat{r}_{\f,t}\right).
\end{align*}
Denote by $Y_{\f,t} = r_{\f,t} - \hat{r}_{\f,t}$. Since
\begin{equation*}
    \E\left[\hat{r}_{\f,t}\bigg|\mathcal{H}_{t}\right] = 
    \begin{cases}
    r_{\f,t} + (1-\epsilon)\alpha\left(1-\mathbb{P}\left(\hat{\f}_{t}= \f |\mathcal{H}_{t}\right)\right)&\mbox{if \, $\f \in \mathcal{U}(\hat{\f}_{t-1}) \cap cond(t)$},\\
   \epsilon r_{\f,t} + (1-\epsilon)\alpha &\mbox{\text{otherwise}},
   \end{cases}
\end{equation*}
and $\alpha > c_{0} \geq r_{\f,t}$ uniformly for any $\f$ and $t$,
then we have uniformly that $\E\left[Y_{t}|\mathcal{H}_{t}\right] \leq 0 $, hence satisfying the first condition. 
\medskip

For the second condition, if $\f \in \mathcal{U}\left(\hat{\f}_{t-1}\right) \cap cond(t)$, then 
\begin{align*}
   \mathrm{Var}(Y_{t}|\mathcal{H}_{t}) =& \E \left[\hat{r}_{\f, t}^{2}|\mathcal{H}_{t}\right] - \left(\E \left[\hat{r}_{\f,t}|\mathcal{H}_{t}\right]\right)^2 \\
    \leq & \epsilon r_{\f,t}^2 + (1-\epsilon)\left[\frac{r_{\f,t}^2}{\mathbb{P}\left(\hat{\f}_{t}= \f |\mathcal{H}_{t}\right)} + \alpha \left(1-\mathbb{P}\left(\hat{\f}_{t}= \f |\mathcal{H}_{t}\right)\right)\right] \\
    &-  \left[r_{\f,t} + (1-\epsilon)\alpha\left(1-\mathbb{P}\left(\hat{\f}_{t}= \f |\mathcal{H}_{t}\right)\right)\right]^2 \\
   \leq & \frac{r_{\f,t}^2}{\beta} + \alpha^{2}(1-\beta)\leq \frac{c_{0}^{2}}{\beta} + \alpha^{2}(1-\beta).
\end{align*}

Similarly, for $\f \not \in \mathcal{U}\left(\hat{\f}_{t-1}\right) \cap cond(t)$, one can have $\mathrm{Var}(Y_{t}|\mathcal{H}_{t}) \leq \alpha^2$.
Moreover, for the third condition, since
\begin{equation*}
    \E \left[Y_{\f,t}|\mathcal{H}_{t}\right] \geq -2\alpha,
\end{equation*}
then 
\begin{equation*}
    Y_{\f, t} - \E \left[Y_{\f,t}|\mathcal{H}_{t}\right] \leq r_{\f,t} + 2\alpha \leq c_{0}+ 2\alpha.
\end{equation*}
Setting $\lambda = \sqrt{2T\left[\frac{c_{0}^2}{\beta}+\alpha^2(1-\beta) +\left(c_{0}+2\alpha\right)^2\right]\ln \left(\frac{1}{\beta}\right)}$ leads to 
\begin{equation*}
    \mathbb{P}\left(\sum_{t=1}^{T}Y_{\f,t} \geq \lambda\right) \leq \beta.
\end{equation*}
Hence the following inequality holds with probability $1-\bigg|\mathcal{F}_{p}\bigg|\beta$
\begin{equation*}
    \max_{\f \in \mathcal{F}_{p}}\sum_{t=1}^{T}\left(r_{\f, t} -\hat{r}_{\f,t}\right) \leq \sqrt{2T\left[\frac{c_{0}^2}{\beta}+\alpha^2(1-\beta) +\left(c_{0}+2\alpha\right)^2\right]\ln \left(\frac{1}{\beta}\right)}.
\end{equation*}
Finally, noticing that $\max_{\f \in \mathcal{F}_{p}}\sum_{t=1}^{T}\left(r_{\f, t} -\hat{r}_{\f,t}\right) \leq c_{0}T$ almost surely, we terminate the proof of \autoref{lem:lemma 6}.

\end{proof}

\begin{proof}[Proof of \autoref{thm:theorem 3}]
Assume that $p >6$, $T \geq 2|\mathcal{F}_{p}|^2$ and let
\begin{multline*}
    \beta = \left|\mathcal{F}_{p}\right|^{-\frac{1}{2}}T^{-\frac{1}{4}}, \qquad \alpha = \frac{c_{0}}{\beta}, \qquad \hat{c}_{0} = \frac{2c_{0}}{\beta}, \\
    \eta_{1} = \eta_{2} =\dots=\eta_{T}= \frac{\sqrt{c_{1}p+c_{2}L+c_{3}}}{\sqrt{T(e-1)}\hat{c}_{0}}, \qquad \epsilon = 1- \left|\mathcal{F}_{p}\right|^{\frac{1}{2} - \frac{3}{p}}T^{-\frac{1}{4}}.
\end{multline*}
With those values, the assumptions of \autoref{lem:lemma 4}, \autoref{lem:lemma 5} and \autoref{lem:lemma 6} are satisfied. Combining their results lead to the following
\begin{align*}
     \cumrwrd \geq & \,    \expglbordonwholepenalty - 2\alpha \beta (1-\epsilon) \sum_{t=1}^{T}\left|\mathcal{U}\left(\hat{\f}_{t-1}\right)\right| \\
     & - \hat{c}_{0}^{2}(e-1)\eta T
      -  \hat{c}_{0}(e-1)\left(c_{1}p+c_{2}L+c_{3}\right) \\
      & - \left(1-\left|\mathcal{F}_{p}\right|\beta\right)\sqrt{2T\left[\frac{c_{0}^2}{\beta}+\alpha^2(1-\beta) +\left(c_{0}+2\alpha\right)^2\right]\ln \left(\frac{1}{\beta}\right)} -  \left|\mathcal{F}_{p}\right|\beta c_{0}T \\
     \geq &  \expglbordonwholepenalty - (1-\epsilon)\left|\mathcal{F}_{p}\right|^{\frac{3}{p}}c_{0}T \\ &- \hat{c}_{0}^{2}(e-1)\eta T 
     - \hat{c}_{0}(e-1)\left(c_{1}p+c_{2}L+c_{3}\right)  \\
     &- \left(1-\left|\mathcal{F}_{p}\right|\beta\right)\sqrt{2T\left[\frac{c_{0}^2}{\beta}+\alpha^2(1-\beta) +\left(c_{0}+2\alpha\right)^2\right]\ln \left(\frac{1}{\beta}\right)} -  \left|\mathcal{F}_{p}\right|\beta c_{0}T \\
     \geq & \expglbordonwholepenalty -  \mathcal{O}\left(\left|\mathcal{F}_{p}\right|^{\frac{1}{2}}T^{\frac{3}{4}}\right),
\end{align*}
where the second inequality is due to the fact that the cardinality $\left|\mathcal{U}\left(\hat{\f}_{t-1}\right)\right|$ is upper bounded by $\left|\mathcal{F}_{p}\right|^{\frac{3}{p}}$ for $t\geq 1$. In addition, using the definition of $r_{\f,t}$ that $r_{\f,t} = c_{0} - \Delta(\f, x_{t})$ terminates the proof of \autoref{thm:theorem 3}.
\end{proof}

\FloatBarrier

\bibliographystyle{plainnat}
\bibliography{biblio}

\end{document}